\documentclass[hidelinks]{article} %
\usepackage[nonatbib,main,preprint]{neurips_2026}

\usepackage[T1]{fontenc}    %
\usepackage{microtype}      %

\usepackage[normalem]{ulem}

\usepackage{xcolor}

\usepackage{mathrsfs}
\usepackage{mathtools,amssymb}   %
\mathtoolsset{showonlyrefs,showmanualtags}

\usepackage{bm}             %
\usepackage{nicefrac}       %

\usepackage{booktabs}

\usepackage[boxed]{algorithm2e}
\SetAlCapHSkip{0pt}
\SetAlCapFnt{\small\rmfamily\centering}
\SetAlCapSkip{0.8ex}
\DontPrintSemicolon            %
\SetAlgoNoLine
\makeatletter
\AtBeginEnvironment{algorithm}{%
  \SetAlgoNoLine %
  \SetAlgoNoEnd
}
\makeatother

\usepackage{my_notation}

\usepackage{amsthm}
\usepackage{thmtools}                   %

\usepackage{my_citations}   %
\usepackage{csquotes}
\usepackage[british]{babel}     %

\usepackage{hyperref}       %
\usepackage{bookmark}           %
\usepackage{zref-clever}

\makeatletter
\let\zc@oldlabel\label
\renewcommand{\label}[1]{\zc@oldlabel{#1}\zlabel{#1}}
\makeatother

\let\cref\zcref
\zcsetup{cap,nameinlink=false}                   %

\AddToHook{env/algorithm/begin}{%
   \zcsetup{reftype=algorithm}}

\zcRefTypeSetup{algorithm}{
Name-sg = Algorithm,
name-sg = algorithm,
Name-pl = Algorithms,
name-pl = algorithms,
}

\usepackage{my_theorems}

\usepackage{enumitem}           %
\usepackage{xparse}

\title{High-probability zeroth-order online convex optimisation beyond Euclidean geometry}

\author{%
  David Janz\\
  University of Oxford\\
  \texttt{david.janz@stats.ox.ac.uk} \\
  \And
  El Mahdi El Mhamdi\\
  École Polytechnique, IP Paris\\
  \texttt{el-mahdi.el-mhamdi@polytechnique.edu}\\
  \And
  Arya Akhavan\\
  University of Oxford\\
  \texttt{arya.akhavan@stats.ox.ac.uk } \\
  }

\begin{document}

\maketitle

\begin{abstract}
We study online convex optimisation with $\ell_q$-Lipschitz losses, $\ell_p$-regularised FTRL, and randomised two-point finite-difference gradient estimators based on cone-measure sampling from $\ell_r$-spheres. For random Lipschitz losses whose mean is convex, we prove unified high-probability regret bounds for all $p,q,r \in [1,\infty]$. The analysis is driven by all-moment bounds for the gradient estimator in the dual FTRL norm, yielding time-uniform control of the quadratic variation. The algorithm is anytime and data-driven; in the special cases previously studied, its rates recover the known in-expectation guarantees while strengthening them to time-uniform high probability. Together with constant-probability lower bounds, these results establish optimality for $q\in[1,2]$ under appropriate sampling geometry, and expose a gap for $q>2$ that appears intrinsic to the estimators themselves.
\end{abstract}

\section{Introduction}
\label{sec:intro}

Zeroth-order optimisation studies settings in which a learner seeks to minimise a function, or a sequence of functions, using only function evaluations \citep{kiefer1952stochastic,blum1954multidimensional,PT90,flaxman2004online,spall2005introduction}. The learner may choose where to evaluate the function, but it does not observe gradients or higher-order information. This restriction arises when gradients are unavailable, unreliable, or too expensive to compute. Although derivative-free optimisation is classical, it has seen renewed
attention in machine learning and statistics, both through bandit and
stochastic convex optimisation models \citep{agarwal2010optimal,shamir2013complexity,duchi2015,nesterov2017random, akhavan2020, Gasnikov,  kornilov2023accelerated, akhavan2024gradient, maoptimal, bashirov2026zeroth}.

This feedback model combines in a nontrivial way with convex optimisation, since convexity is powerful precisely because first-order information is global: a subgradient gives a supporting hyperplane, and hence a direction in which progress can be made. In this sense, convex optimisation almost begs for gradient-based methods. The standard way to proceed in zeroth-order convex optimisation is therefore to use a number of function evaluations to estimate a gradient---as in finite differences---and then use this estimate with a first-order algorithm such as gradient/mirror descent or follow-the-regularised-leader (FTRL) \citep{flaxman2004online,duchi2015,akhavan2022gradient}. The guarantees of this reduction depend on how the variation of the objective controls the finite-difference estimate, and how large the estimate is in the norm seen by the first-order algorithm.

We study this problem for target functions that are Lipschitz with respect to an $\ell_q$ norm, $q \in [1,\infty]$, and for FTRL regularisers that are strongly convex with respect to an $\ell_p$ norm, $p \in [1,\infty]$. We analyse  randomised two-point gradient estimators where the target function is queried on either side of the current iterate along a direction chosen according to the cone measure on the boundary of an $\ell_r$ ball, $r \in [1,\infty]$. Here, taking $r=2$ yields a classical estimator, whereas \(r=1\) can improve the dimension dependence for some problem geometries. Our goal is to prove high-probability regret bounds for this family of geometries. This requires tail control for the gradient estimator in the $\ell_{p^*}$-norm (where $p^*$ denotes the H\"older conjugate of $p$), which in our analysis comes from controlling all moments; in contrast,  in-expectation analysis uses at most the second moment \citep[as in, e.g.,][]{akhavan2022gradient}.

Our contributions are as follows.
\begin{enumerate}
    \item We prove moment bounds for $\ell_r$-randomised two-point gradient estimators in the dual norm $\ell_{p^*}$, tracking the dependence on the $\ell_r$ sampling geometry and the $\ell_q$ Lipschitz norm. These imply concentration bounds for the cumulative squared dual norms that appear in FTRL analyses. The estimator-level results are not specific to FTRL and may be useful in other analyses of derivative-free methods based on randomised finite differences.

    \item We combine these estimators with an adaptive FTRL algorithm and prove high-probability regret bounds for random Lipschitz losses with a convex mean. The algorithm is anytime and data-driven: it does not require as input the horizon or any worst-case quantity from the moment analysis (it does need a choice of regulariser and upper-bound on the domain size).

    \item We prove algorithm-independent constant-probability lower bounds for the same problem class. The lower bounds improve on previous in-expectation results for $p=q=1$, and highlight that standard randomised gradient estimators may be suboptimal when $q > 2$.
\end{enumerate}
To our knowledge, our results give the first unified treatment of geometries $p,q, r \in [1,\infty]$, and the first high-probability analysis for the problem  beyond the Euclidean setting ($p=q=r=2$).

\section{Problem setup and main results}

This paper considers the following online convex optimisation problem. Let $\mu$ be a distribution over  real-valued functions on $\Rd$ such that each $f$ in the support of $\mu$ is $L$-Lipschitz with respect to $\|\cdot\|_q$, and the mean function
\[
F_\mu(x) = \E_{f \sim \mu}[f(x)]
\]
is convex. Observe that we do not require realisations $f \sim \mu$ themselves to be convex. Let $f_1,f_2,\dots$ be independent random functions distributed according to $\mu$, and fix a compact, convex set of comparators $\Theta \subset \Rd$. At each round $t=1,2,\dots$, a learner picks an estimate $x_t \in \Theta$ of the minimum of $F_\mu$, and is then allowed to query $f_t$ at exactly two locations. The learner receives the feedback for both queries simultaneously; the second query cannot depend on the first feedback.

\begin{remark}[Lipschitz $F_\mu$ versus $f$]
  The closest in-expectation analysis in this setting assumes only that the mean function $F_\mu$ is Lipschitz \citep{duchi2015}. We impose the stronger condition that the samples $f \sim \mu$ are themselves Lipschitz to allow for high-probability control of the regret.
\end{remark}

We judge the performance of the learner according to the $n$-step (pseudo)regret incurred, defined as
\[
  R_n^\mu = \sup_{u\in \Theta}\sum_{t=1}^n \roundb[\big]{F_\mu(x_t) - F_\mu(u)}\,.
\]
Since $x_1,x_2,\dots$ depend on the observed zeroth-order feedback, the regret is a random variable.

\begin{remark}[Relation to bandit convex optimisation]
Our optimisation setting allows the two function queries at each round to
be made anywhere in \(\mathbb R^d\). This differs from bandit convex
optimisation, where the queried action is itself the learner's decision
and must lie in the constraint set. The bandit constraint makes boundary
effects and feasible exploration central issues; see \citet{flaxman2004online,agarwal2010optimal} for early formulations and
\citet{lattimore2024bandit} for a recent survey.
\end{remark}

The algorithm analysed below uses the two function evaluations at round \(t\) to form a random vector \(g_t\), which plays the role of a gradient estimate at the current iterate \(x_t\). These estimates are then passed to an adaptive FTRL update. The regret depends on the cumulative squared dual norms
\[
    Y_n=\sum_{t=1}^n \|g_t\|_{p^*}^2,
\]
where \(\|\cdot\|_{p^*}\) is the dual norm to the \(\ell_p\)-geometry used by the regulariser; this is standard in the analyses of FTRL. However, in the zeroth-order setting, we will also have to account for the error introduced by having the algorithm run on gradient estimates, rather than true first-order information.

We now define the \(\ell_r\)-cone-measure gradient estimates, state the adaptive FTRL procedure, give its high-probability regret guarantees, and present constant-probability lower bounds.

\subsection{Randomised cone measure two-point gradient estimates}

Let $K$ be a convex symmetric body in $\Rd$. The cone measure $\mu_{K}$ on the surface $\partial K$ of the body $K$ is induced by taking a random variable uniformly in $K$, and then applying a radial projection to~$\partial K$. Formally, for $A \subset \partial K$,
\[
  \mu_K(A) = \frac{\vol\{ ta \colon a \in A,\, 0 \leq t \leq 1\}}{\vol K}\,.
\]
In this paper, we will take the convex body $K$ to be an $\ell_r$ ball of the form $B^d_r = \{x \in \Rd \colon \norm{x}_r \leq 1\}$. We will write $\zeta \sim \partial B^d_r$ to denote that $\zeta$ is distributed according to the cone measure $\mu_{B^d_r}$.

The gradient estimators we consider will also need the notion of a normal to the surface of an $\ell_r$ ball.

\begin{definition}[Unit normal]\label{def:unit-normal}
Let \(1\le r\le\infty\). We define the unit normal to \(B_r^d\) in the
direction \(\zeta\in\R^d\) as
\[
n_r(\zeta)\in\argmax_{x\in B_{r^*}^d}\langle x,\zeta\rangle.
\]
Equivalently, letting \(r^*\) denote the H\"older conjugate of \(r\), \(n_r(\zeta)\) is any vector satisfying
\[
\norm{n_r(\zeta)}_{r^*}=1,
\qquad
\langle n_r(\zeta),\zeta\rangle=\norm{\zeta}_r\,.
\]
\end{definition}

For concreteness, these take the following closed forms.

\begin{lemma}[name=Closed-form normals,restate=LemmaClosedFormNormal]\label{claim:normal}For $r \geq 1$ and $\zeta \in \partial B^d_r$, the normals admit the closed forms
\[
  n_1(\zeta)=\sign(\zeta),
  \qquad
  n_r(\zeta)_i=\sign(\zeta_i)|\zeta_i|^{r-1},
  \quad 1<r<\infty\,.
\]
If \(r=\infty\), let \(J\in\argmax_{i\le d}|\zeta_i|\). Then we may take $n_\infty(\zeta)=\sign(\zeta_{J})e_{J}$.
\end{lemma}

Here and throughout, we adopt the convention that \(\sign(0)=1\). Beware, as our unit normal has unit $\ell_{r^*}$ norm; you may be more familiar with the common convention that the normal has unit $\ell_2$ norm.

The gradient estimates we consider throughout this paper are constructed as follows.

\begin{definition}[$\ell_r$-randomised gradient estimate]
    Fix $r \in [1,\infty]$ and let $\zeta \sim \partial B^d_r$. Let $h > 0$ be a width parameter. We consider the gradient estimates given by
    \[
    g_h(x,\zeta; f) = \Delta_h(x,\zeta; f)n_r(\zeta) \spaced{where}
    \Delta_h(x,\zeta; f) = \frac{d}{2h}\roundb[\big]{f(x+h\zeta) - f(x-h\zeta)}
  \]
  is the two-point gradient magnitude estimate and $n_r(\zeta)$ denotes the normal to the $\ell_r$ ball surface at $\zeta$.
\end{definition}

\subsection{Algorithm and upper bound}

The algorithm is SOLO FTRL \citep{orabona2018scale}, run on $\ell_r$-randomised gradient estimates.

\begin{definition}[Algorithm]\label{def:alg}
    Fix a regulariser $V \colon \Theta \to \R$ and let the starting iterate be the minimiser \[
    x_1 = \argmin_{u \in \Theta} V(u)\,.
  \] At round $t=1,2,\dots$, choose a width $h_t > 0$, sample \(\zeta_t\) from the cone measure on \(\partial B_r^d\) and query $f_{t}$ at $x_t \pm h_t\zeta_{t}$. Use the feedback to compute the gradient estimate $g_{t} := g_{h_t}(x_t, \zeta_t; f_t)$.
Choose the next iterate according to some fixed measurable selection of the argmin correspondence
\[
x_{t+1} \in \argmin_{u \in \Theta} \curlyb[\Bigg]{ \Bigl\langle \sum_{s=1}^t g_s,\, u \Bigr\rangle + \frac{V(u)}{R} \sqrt{2.75 \cdot Y_t}}\,, \qquad Y_t = \sum_{s=1}^t \norm{g_s}_{p^*}^2\,.
\]
\end{definition}

For the analysis, we require the following assumptions on the domain $\Theta$ and the regulariser $V$.

\begin{assumption}[Regulariser]\label{ass:reg}
  Let $\Theta \subset \mathbb R^d$ be compact and convex and $V \colon \Theta \to \R$ be nonnegative, lower semicontinuous and $1$-strongly convex with respect to $\norm{\cdot}_p$. Define $R^2 = \sup_{u \in \Theta} V(u)$.
\end{assumption}

That the regulariser is nonnegative is for convenience only; any regulariser can be simply offset.

Our upper bound will use the constants $C_r,c_r > 0$ defined in the following concentration inequality.

\begin{theorem}[name=Concentration of Lipschitz functions under the cone measure,restate=lipschitzconcentration]\label{thm:lipschitz-concentration-1}
  Fix $r \in [1,\infty)$ and $\zeta \sim \partial B_r^d$. There exist constants $C_r, c_r > 0$ independent of $d$, such that for every $1$-Lipschitz function $\phi$ on $(\partial B^d_r, \norm{\cdot}_2)$ and every $u > 0$,
  \begin{align}
    \P{|\phi(\zeta) - \E \phi(\zeta)| > u } \leq C_r \exp\{-  d^{\frac{2}{r \vee 2}} u^{r \wedge 2}/c_r\}\,.
  \end{align}
\end{theorem}

The inequality is a generalisation of L\'evy's spherical concentration inequality (case $r=2$, where $C_2=2$ and $c_2=4$ are admissible constants). It was established for $r \in [1,2)$ by \citet{schechtman2000concentration} and extended to $r > 2$ by \citet{naor2007surface}. A limitation of our result is that for $r\neq2$ we do not have explicit constants $C_r,c_r$. The cited proofs can in principle yield constants, but they are not numerically informative; in any case, the algorithm is specifically designed not to need them.

We now define the variance proxy, the square root of which gives the leading factor of the regret, and which describes the typical scale of the squared variation of the dual norm of the gradient estimates.

\begin{definition}[name=Proxy variance $\nu_r$,restate=defVar]\label{def:variance}
  Define
  \[
  K_{p,r} = 4C_r^{1/4} c_r^{1/(r \wedge 2)}
  \begin{cases}
    (10r)^{1/r^*} & p=1\\
    (1+r)^{1/4}\roundb{1+(r-1)(p^*\vee4)}^{1/r^*} & p \in (1,\infty]
  \end{cases}\,,
\]
which is a constant depending on $p,r$ only, as well as the $p=1$ suboptimality term
\[
  D_{p,r}(d) = \begin{cases}
    (\log d)^{1/r^*} & p=1 \\
    1 & p \in (1,\infty]\\
  \end{cases}\,.
\]
The proxy variance $\nu_r >0$, which will feature in the leading term of the upper bound, is given by
\[
  \sqrt{\nu_r} = L K_{p,r} D_{p,r}(d) d^{1/p^*+(1/q - 1/2)_+} \,, \qquad r \in [1,\infty)\,, \spaced{and} \sqrt{\nu_\infty} = \sqrt{2} L d^{1+(1/q - 1/2)_+}\,.
\]
\end{definition}

Establishing that $\nu_r$ is a proxy variance for the dual gradient norms in a suitable Bernstein moment condition/subgamma moment generating function sense is one of the core contributions of this work.

\begin{theorem}[Upper bound]\label{thm:upper-bound}
  Fix $d \geq 2$, $p,q, r \in [1,\infty]$. Suppose $x_1,x_2,\dots \in \Theta$ are the iterates generated by the algorithm of \cref{def:alg}, with widths satisfying $\sum_{t=1}^\infty h_t \leq \frac{R}{d}$ and the domain $\Theta$ and regulariser $V$ satisfying \cref{ass:reg}. Recall $R^2 = \sup_{u \in \Theta} V(u)$. Write
  \[
  \gamma_r = (1/p-1/r)_+\,, \qquad   r \in [1,\infty) \spaced{and} \gamma_\infty = 1/(q \vee 2)\,.
  \]
  Then, for any $\delta \in (0,1)$ and any problem instance $\mu$,
\[
    \P[\bigg]{\forall n \geq 1 \colon  R_n^\mu \leq  22R\sqrt{\nu_r}\roundb[\Big]{\sqrt{n\Lambda^\star_n(\delta)} + 11d^{\gamma_r} \log(32n/\delta)^{3/2}}} \geq 1-\delta\,,
  \]
  where $\Lambda_n^\star(\delta) = 2\log\log(e^e + n) + \log(4/\delta)$.
\end{theorem}

The algorithm only needs the specification of the choice of the geometries $p,r \in [1,\infty]$, the regulariser $V$, the value of $R^2 = \sup_{u\in \Theta} V(u)$ (or an upper bound on $R^2$), and the dimension $d$ of the problem. Beyond that, it is fully parameter free and adaptive: none of the constants defined thus far matter for implementation. The $\sqrt{\log\log n}$ in the leading term of the regret is due to the anytime nature of the bound. For every fixed $n$, the bound holds without that factor after adjusting constants.

\subsection{Lower bound}\label{sec:lower-bound}

Our high-probability upper bound should be contrasted with this constant-probability lower bound.

\begin{theorem}[name=Lower bounds,restate=lowerbound]
\label{thm:lower-bound}
Fix $2 \leq d \leq n$, \(p,q\in[1,\infty]\) and \(R,L>0\). Let
\[
    \Theta = R B_p^d=\curlyb{x\in\Rd\colon \norm{x}_p\le R}.
\]
Let \(\cM\) be the class of probability laws \(\mu\) on real-valued functions on
\(\Rd\) such that, if \(f\sim\mu\), then \(f\) is convex and \(L\)-Lipschitz
with respect to \(\norm{\cdot}_q\). For a two-point algorithm
\(\cA\), let \(\mathbf P_\mu^\cA\) denote the probability law induced by running
\(\cA\) with independent losses \(f_1,\dots,f_n\sim\mu\). Then
\[
    \sup_{\mu\in\cM}
    \mathbf P_\mu^\cA
    \curlyb[\bigg]{
        R_n^\mu
        \ge
        d^{(1/2+1/q-1/p)_+}\frac{LR\sqrt{n}}{256}
    }
    \ge
    \frac17\,.
\]
Moreover, letting $x_n$ denote the output of algorithm $\cA$ at round $n$, the optimisation error satisfies
\[
    \sup_{\mu\in\cM}
    \mathbf P_\mu^\cA
    \curlyb[\bigg]{
        F_\mu(x_n)-\inf_{x\in\Theta}F_\mu(x)
        \ge
        d^{(1/2+1/q-1/p)_+} \frac{LR}{256\sqrt{n}}
    }
    \ge
    \frac17\,.
\]
\end{theorem}

\paragraph{Comparing upper and lower bounds} Comparing our algorithm-specific upper bound to this algorithm-agnostic lower bound shows the bounds are generally tight, except for three settings:
\begin{enumerate}
  \item For the case $p=1$, the upper bound has an extra $(\log d)^{1/r^*}$ factor. This might not be an artefact of the proof: choosing $r=1$ yields the optimal dependence, and it is entirely plausible that $\ell_1$-cone sampling works better on domains that are small in $\ell_1$ geometry. \citet{akhavan2022gradient} previously argued that $\ell_1$ randomisation is effective on the simplex.

  \item The algorithm's upper bound does not match the algorithm-independent lower bound in the regime $q > 2$, which was not studied in the in-expectation results of \citet{duchi2015}. We conclude in \cref{sec:q-suboptimal} that this looks to be a suboptimality of the $\ell_r$-randomised two-point gradient estimator itself, rather than a weakness of our analysis or lower bound.

  \item Taking \(r=\infty\) in the algorithm construction also gives an
  upper bound that fails to match the lower bound. As verified in
  \cref{prop:cube-endpoint-tight}, our estimate of \(\nu_\infty\) is tight, so this does not look
  like a proof artefact. The more natural conclusion is that the
  \(\ell_\infty\)-cone-measure gradient estimator is simply not the right tool:
  its closed form, shown in \cref{lem:cube-cone-representation}, looks odd enough
  that one ought not be too surprised if and when it performs poorly.
\end{enumerate}

\paragraph{On the lower bound} The closest relevant lower-bound result in the literature is due to \citet{duchi2015}, which covers the cases \(q=1\) and \(q=2\) with \(p\in[1,\infty]\), but gives an in-expectation result rather than a constant-probability result. In the Euclidean Lipschitz case \(q=2\), our lower bound matches the
in-expectation lower bound of \citet{duchi2015} for all
\(p\in[1,\infty]\). In the case \(p=q=1\), our lower bound improves the
dimension dependence of their in-expectation lower bound by removing a factor of
\(\sqrt{\log d}\). Our result is the first to cover the full range \(p,q \in [1,\infty]\) jointly, and we do so with a single construction. Our construction differs significantly from \citet{duchi2015}:
\begin{enumerate}
  \item Since our upper bound assumes the functions $f \sim \mu$ are almost surely Lipschitz, we require our lower bound to be realised through measures supported on Lipschitz functions. This rules out the family of Gaussian measures over weights of linear functions used by \citet{duchi2015}. Instead, we use shifted versions of the classic compactly supported biweight kernel to define the density of our measures \citep[page 3]{tsybakov2009nonparametric}. Compact support yields the $\sqrt{\log d}$ improvement for the $p=1$ case.
  \item We also optimise over the dimension $s \leq d$ of the support of the hard cases, picking $s < d$ when $q > 2$. This optimisation prevents the lower bound from shrinking with the ambient dimension \(d\), and highlights that the $q > 2$ regime is intrinsically different from $q \in [1,2]$.
\end{enumerate}
There is a question of whether our lower bound is tight in the $q > 2$ regime; we explore this further in \cref{sec:q-suboptimal}. Our lower bounds are established in \cref{appendix:lower-bounds}.

\section{Proof of the upper bound}

Our upper bound hinges on establishing that the sequence of dual norm moments of the gradient estimates satisfy the following sequential condition, which implies that it is a subgamma process.

\begin{definition}[Sequential Bernstein moments]
  Let $(\FF_t)_{t \geq 0}$ be a filtration and $\nu,c$ positive real numbers. An adapted real-valued process $(X_t)_{t \geq 1}$ satisfies the sequential Bernstein moment condition $\SBM$ if, for every $t \geq 1$ and every integer $k \geq 2$,
  \[
    \E[|X_t|^k \mid \FF_{t-1}] \leq \frac{k!}{2}\nu c^{k-2} \qquad a.s.
  \]

\end{definition}

In particular, with $(\FF_t)_{t\in\N}$ being the filtration generated by $(g_t)_{t \geq 1}$, the following holds.

\begin{proposition}[name=,restate=propGradBound]\label{lem:gradient-moment-bound}
Let \(d\ge2\), \(p,q,r\in[1,\infty]\) and assume the setting of \cref{thm:upper-bound} holds. Write $\gamma_r = (1/p - 1/r)_+$ for $r \in [1,\infty)$ and $\gamma_\infty = 1/(q \vee 2)$. The process \[
  \norm{g_1}_{p^*}, \norm{g_2}_{p^*}, \dots \spaced{satisfies} \operatorname{SBM}(\nu_r, d^{\gamma_r} \sqrt{\nu_r})
\] with respect to the filtration $(\FF_t)_{t \geq 1}$ for the earlier stated variance proxy $\nu_r$ given in \cref{def:variance}.
\end{proposition}

We first show how this moment bound implies \cref{thm:upper-bound}. We then outline the proof of the moment bound itself; the full proof is deferred to \cref{appendix:grad-bounds}.

\subsection{Proof of \cref{thm:upper-bound}, regret upper bound}

The following proposition isolates how regret depends on the randomised gradient estimates.

\begin{proposition}[name=Regret bound,restate=thmRegret]\label{prop:regret-decomp}
  Select a sequence of positive widths $(h_t)_{t \geq 1}$ such that $\sum_{t=1}^\infty h_t \leq \frac{R}{d}$. Let $u^\star \in \argmin_{u \in \Theta} F(u)$ and define
  \[
    M_n = \sum_{t=1}^n \langle g_t - \E_{t-1}[g_t], u^\star - x_t \rangle\,, \qquad Y_n = \sum_{t=1}^n \norm{g_t}_{p^*}^2\,.
  \]
  Then, for every $n \geq 1$,
 \begin{align*}
        R_n \leq LR  + 13.3R\sqrt{Y_n} + M_n\,.
 \end{align*}
\end{proposition}

The proof will rely on standard results relating randomised gradient estimators and smoothing.

\begin{lemma}\label{lem:smoothing}
  Let $\phi \colon \Rd \to \R$ be $L$-Lipschitz with respect to $\norm{\cdot}_q$, $h > 0$ and $U \sim B^d_r$. Define the smoothed function
  \[
    \phi_h(x) = \E[\phi(x+hU)]\,.
  \]
  Then the smoothed function $\phi_h$ is differentiable, with derivative given by
  \[
    \nabla \phi_h(x) = \E[ \frac{d}{2h} \roundb{\phi(x+h\zeta) - \phi(x-h\zeta)} n_r(\zeta) ]\,, \qquad \zeta \sim \partial B^d_r\,.
  \]
  Moreover, if $\phi$ is convex, then so is $\phi_h$, and the bias of smoothing is controlled, in that
  \[
    0 \leq \phi_h(x) - \phi(x) \leq Lh \E \norm{U}_q\,.
  \]
\end{lemma}

The derivative identity is the Gauss--Green formula for the average of
$\phi$ over $x+hB_r^d$, rewritten using the cone measure on $\smash{\partial B_r^d}$,
and extended to Lipschitz functions by mollification. For $r=2$ this is the
classical spherical smoothing identity appearing in
\citet[Chapter~9, Section~9.3.2, Exercise~2]{nemirovskij1983problem} and used
in zeroth-order online optimisation by \citet{flaxman2004online}. The
$\ell_1$ version is proved by \citet{akhavan2022gradient}; the same argument gives the stated form for all $r\ge1$. For \(r=\infty\), the same identity follows by applying the divergence theorem face by face on the cube; the intersections of faces have zero surface measure.

\begin{proof}[Proof of \cref{prop:regret-decomp}]
  Write $F:=F_\mu$. Observe that $F$ is $L$-Lipschitz with respect to $\norm{\cdot}_q$. Indeed, writing $f \sim \mu$, for every $x,x' \in \R^d$,
  \[
    |F(x) - F(x')| = |\E [f(x) - f(x')]| \leq L\norm{x-x'}_q\,.
  \]

  Let $F_{h_t}$ denote the $h_t$-smoothing of the mean function $F$. By the bias control property of \cref{lem:smoothing},
  \[
    R_n = \sum_{t=1}^n \roundb[\big]{F(x_t) - F(u^\star)} \leq L \E \norm{U}_q \sum_{t=1}^n h_t + \sum_{t=1}^n \roundb{ F_{h_t}(x_t) - F_{h_t}(u^\star)}\,.
  \]
  Here, note that by norm equivalence and since $U \in B^d_r$,
  \[
    \E \norm{U}_q \leq d^{(1/q-1/r)_+} \E \norm{U}_r \leq d^{(1/q-1/r)_+} \leq d\,,
  \] so our choice of $(h_t)_{t \geq 1}$ means the first term cannot exceed $LR$. Consider the remaining sum. Since each $F_{h_t}$ is differentiable and convex (again by \cref{lem:smoothing}),
  \[
    \sum_{t=1}^n \roundb{ F_{h_t}(x_t) - F_{h_t}(u^\star)} \leq \sum_{t=1}^n \langle \nabla F_{h_t}(x_t), x_t - u^\star \rangle\,.
  \]
  And adding-and-subtracting, we obtain that
  \[
      \sum_{t=1}^n \langle \nabla F_{h_t}(x_t), x_t - u^\star \rangle = \sum_{t=1}^n \langle g_t, x_t - u^\star \rangle - \sum_{t=1}^n \langle g_t - \nabla F_{h_t}(x_t), x_t - u^\star \rangle\,.
  \]
  By the guarantee for SOLO FTRL on bounded domains, Theorem 1 of \citet{orabona2018scale},
  \[
    \sum_{t=1}^n \langle g_t, x_t - u^\star \rangle \leq 13.3R\sqrt{ Y_n}\,.
  \]
  It remains to establish that $\nabla F_{h_t}(x_t) = \E_{t-1}[g_t]$, identifying the additive term with $M_n$.

  By the derivative identity of \cref{lem:smoothing}, and since by definition, $F(x) = \E[f(x)]$, we have that
  \begin{align*}
        \nabla F_{h_t}(x_t)
        &= \E_\zeta[\frac{d}{2h_t}\roundb{F(x_t + h_t\zeta) - F(x_t - h_t\zeta)} n_r(\zeta)] \\
        &= \E_{\zeta,f}[\frac{d}{2h_t}\roundb{f(x_t + h_t \zeta) - f(x_t - h_t\zeta)} n_r(\zeta)] \\
        &= \E_{t-1}[\frac{d}{2h_t}\roundb{f_t(x_t + h_t \zeta_t) - f_t(x_t - h_t\zeta_t)} n_r(\zeta_t)] = \E_{t-1}[g_t]\,,
  \end{align*}
  where $\E_\zeta$ and $\E_{\zeta,f}$ denote the expectations over $\zeta$ and $(\zeta, f)$, respectively; to change from $\E_{\zeta,f}$ to $\E_{t-1}$ note $x_t$ is $\FF_{t-1}$-measurable, $\zeta_t \sim \partial B^d_r$ and $f_t \sim \mu$, and $\sigma(\zeta_t,f_t)$ is independent of~$\FF_{t-1}$.
\end{proof}

We now control $Y_n$ and $M_n$, result stated in the following proposition. Given the proposition, the proof of \cref{thm:upper-bound} is just a straightforward calculation, given immediately thereafter.

\begin{proposition}[Control of $Y_n$ and $M_n$]\label{prop:YM-control}
  With probability at least $1-\delta$, for all $n \geq 1$,
\[
  \sqrt{Y_n} \leq \sqrt{\nu_r n} + 3\sqrt{\nu_r} \log(32n^2/\delta) + d^{\gamma_r}\sqrt{\nu_r}\log(32n^2/\delta)^{3/2}
\]
and
\[
  |M_n| \leq 3R\sqrt{8n\nu_r\Lambda_n^\star(\delta)}
    +10\sqrt{8\nu_r}Rd^{\gamma_r}\Lambda_n^\star(\delta)\,.
\]
\end{proposition}

\begin{proof}[Proof of \cref{thm:upper-bound}]
  On the event where the above inequalities hold, from \cref{prop:regret-decomp} and by bounding terms through the crude inequality $1 \leq \Lambda^\star_n(\delta) \leq d^{\gamma_r}\log(32n^2/\delta)$, we obtain for all $n \geq 1$,
\[
  R_n \leq RL + R\sqrt{\nu_r} \roundb[\big]{22\sqrt{n\Lambda_n^\star(\delta)} + 82 d^{\gamma_r}\log(32n^2/\delta)^{3/2}}\,.
\]
The main theorem follows by absorbing $L$ into the lower order term.
\end{proof}

It remains to prove \cref{prop:YM-control}.  We will use the following two concentration results.

\begin{theorem}[name=Weibull-type process concentration,restate=weibullTheorem]\label{thm:weibul-time-uniform}
Fix $\delta \in \roundb{0,1}$. Let $(\bA_t)_{t \geq 0}$ be a filtration and $(X_t)_{t \geq 1}$ a nonnegative adapted process satisfying $\SBM(\nu,c)$. Then with probability at least $1-\delta$, for all $n \geq 1$,
\[
  \squareb[\bigg]{\sum_{t=1}^n X_t^2}^{\frac{1}{2}} \leq \sqrt{\nu n} + 3\sqrt{\nu}\log(16n^2/\delta) + c\roundb{\log(16n^2/\delta)}^{3/2}\,.
\]
\end{theorem}

\begin{theorem}[name=Bernstein condition process concentration,restate=subgammaconcentration]\label{thm:bernstein-process-concentration}
Fix $\delta \in \roundb{0,1}$. Let $(\bA_t)_{t \geq 0}$ be a filtration and $(X_t)_{t \geq 1}$ an adapted process satisfying $\SBM(\nu,c)$. Then with probability at least $1-\delta$, for all $n \geq 1$
\[
    \abs[\Big]{\sum_{t=1}^n \roundb[\big]{X_t - \E[ X_t \mid \bA_{t-1}]}} < 3\sqrt{n \nu \Lambda_n(\delta)} + 10c \Lambda_n(\delta)\,\,
\]
where $\Lambda_n(\delta) = 2\log\log(e^e + n\nu/c^2) + \log(2/\delta)$.
\end{theorem}

Both theorems are proven in \cref{appendix:concentration}. The summands in the first theorem are heavy-tailed, and so the proof combines a truncation argument and Bernstein's inequality to deal with the probability of large jumps, and Freedman's inequality applied to the truncated process. The second result, a time-uniform Bernstein-type inequality, is an application of Theorem 3.1 in \citet{whitehouse2023time}.

\begin{proof}[Proof of \cref{prop:YM-control}]
  By \cref{lem:gradient-moment-bound}, the process
  \((\norm{g_t}_{p^*})_{t \geq 1}\) satisfies \(\SBM(\nu_r,  d^{\gamma_r}\sqrt{\nu_r})\) with respect to $(\FF_t)_{t\geq 1}$. Hence, applying
  \cref{thm:weibul-time-uniform} with confidence level \(\delta/2\), we have
  with probability at least \(1-\delta/2\), for all \(n\geq1\), \[
    \sqrt{Y_n} = \roundb[\Big]{\sum_{t=1}^n \norm{g_t}_{p^*}^2}^{1/2} \leq \sqrt{\nu_r n} + 3\sqrt{\nu_r} \log(32n^2/\delta) + d^{\gamma_r}\sqrt{\nu_r}(\log(32n^2/\delta))^{3/2}\,.
  \]
  To control \(M_n\), first note that nonnegativity and strong convexity of \(V\) imply that for all $x,u \in \Theta$,
  \[
    0 \leq V\roundb{\frac{x+u}{2}}
    \leq \frac{V(x)+V(u)}{2} - \frac{1}{8}\norm{x-u}_p^2,
  \]
  so, since \(0\leq V\leq R^2\) on \(\Theta\),
  \[
    \norm{x-u}_p^2 \leq 4V(x)+4V(u)\leq 8R^2.
  \]
  Now let \(\xi_t=\langle g_t,u^\star-x_t\rangle\). Since \(x_t-u^\star\) is \(\FF_{t-1}\)-measurable, H\"older's inequality and the above bound on $\norm{x_t - u^\star}_p$ (with $x_t,u^\star \in \Theta$), give that for every integer \(k\geq2\),
  \[
    \E_{t-1}|\xi_t|^k
    \leq \norm{x_t-u^\star}_p^k\E_{t-1}\norm{g_t}_{p^*}^k
    \leq (\sqrt{8}R)^k\frac{k!}{2}\nu_r (d^{\gamma_r} \sqrt{\nu_r})^{k-2}
    =
    \frac{k!}{2}(8R^2\nu_r)(\sqrt{8\nu_r}Rd^{\gamma_r})^{k-2}.
  \]
  Therefore, \((\xi_t)_{t\geq1}\) satisfies $\operatorname{SBM}(8R^2\nu_r,\sqrt{8\nu_r}Rd^{\gamma_r})$. Applying
  \cref{thm:bernstein-process-concentration} with confidence level
  \(\delta/2\), we obtain that with probability at least \(1-\delta/2\), for all \(n\geq1\),
  \[
    \abs[\Big]{\sum_{t=1}^n\roundb{\xi_t-\E_{t-1}\xi_t}}
    \leq
    3R\sqrt{8n\nu_r\Lambda_n(\delta/2)}
    +10\sqrt{8\nu_r}Rd^{\gamma_r}\Lambda_n(\delta/2).
  \]
  Using that $\frac{8R^2\nu_r}{(\sqrt{8\nu_r}Rd^{\gamma_r})^2} \leq 1$, we have $\Lambda_n(\delta/2) \leq \Lambda^\star_n(\delta)$. And since \(x_t-u^\star\) is \(\FF_{t-1}\)-measurable,
  \[
    \sum_{t=1}^n\roundb{\xi_t-\E_{t-1}\xi_t}
    =
    \sum_{t=1}^n\langle g_t-\E_{t-1}g_t,u^\star-x_t\rangle
    =
    M_n.
  \]
  The result follows by the union bound over the two time-uniform events.
\end{proof}

\subsection{Overview of the proof of \cref{lem:gradient-moment-bound}, gradient moment bound}\label{sec:moment-control}

We begin by reducing the claim to an unconditional moment estimate. Fix a round
\(t\), and condition on \(\FF_{t-1}\). Then \(x_t\) and \(h_t\) are fixed,
while \(f_t\sim\mu\) and \(\zeta_t\sim\partial B_r^d\) are sampled independently
of the past. The estimates below are uniform over the realised
\(L\)-Lipschitz function \(f_t\), so it is enough to consider fixed
\(x\in\Rd\), \(h>0\), and a fixed \(L\)-Lipschitz function \(f\), with
randomness only in \(\zeta\sim\partial B_r^d\). Write
\[
  \Delta(\zeta)
  =
  \frac{d}{2h}\roundb[\big]{f(x+h\zeta)-f(x-h\zeta)} \spaced{and} g(\zeta)=\Delta(\zeta)n_r(\zeta)\,,
\]
such that the quantity we are interested in bounding is $\E \norm{g(\zeta)}_{p^*}^k = \E[|\Delta(\zeta)|^k \norm{n_r(\zeta)}_{p^*}^k]$ for $k \geq 2$.

We begin by applying a worst-case bound on all but two of the powers of $\norm{n_r(\zeta)}_{p^*}$, so that
\[
  \E \norm{g(\zeta)}_{p^*}^k \leq \sup_{\xi \in \partial B^d_r} \norm{n_r(\xi)}_{p^*}^{k-2} \E[|\Delta(\zeta)|^k \norm{n_r(\zeta)}_{p^*}^2]\,.
\]
This worst-case bound is harmless, as terms raised to the power $k-2$ affect only the scale term $c$, which is lower order in the regret bound. Since $n_r(\xi) \in B^d_{r^*}$ for all $\xi \neq 0$, by norm equivalence
\[
  \sup_{\xi \in \partial B^d_r} \norm{n_r(\xi)}_{p^*} \leq d^{(1/p^*-1/r^*)_+} \sup_{\xi \in B^d_r} \norm{n_r(\xi)}_{r^*} = d^{(1/r-1/p)_+}\,.
\]
(The expectation $\E \norm{n_r(\zeta)}_{p^*}$ scales with $d^{1/r-1/p}$, but using it complicates the proof.) The natural next step is to separate the two distinct remaining random quantities via Cauchy-Schwarz, bounding
\[
   \E[|\Delta(\zeta)|^k \norm{n_r(\zeta)}_{p^*}^2] \leq \sqrt{\E[|\Delta(\zeta)|^{2k}]} \sqrt{\E[\norm{n_r(\zeta)}_{p^*}^4]}\,.
\]

The finite-difference term is controlled by concentration of Lipschitz functions
on the \(\ell_r\) sphere. Indeed, for any \(\zeta,\zeta'\in\partial B_r^d\), by the Lipschitz property of $f$ and norm equivalence on $\Rd$,
\[
\begin{aligned}
  |\Delta(\zeta)-\Delta(\zeta')|
  &\leq
  \frac{d}{2h}\roundb[\big]{
      |f(x+h\zeta)-f(x+h\zeta')|
      +
      |f(x-h\zeta)-f(x-h\zeta')|
  }  \\
  &\leq
  dL\norm{\zeta-\zeta'}_q
  \leq
  d^{1+(1/q-1/2)_+}L\norm{\zeta-\zeta'}_2 .
\end{aligned}
\]
The parameter \(h\) cancels, as it should. Moreover, \(\Delta\) is odd and the cone measure is symmetric, so \(\E\Delta(\zeta)=0\). Applying \cref{thm:lipschitz-concentration-1} gives the
Weibull tail
\[
  \P[\big]{|\Delta(\zeta)|>u}
  \leq
  C_r\exp\curlyb[\big]{-(u/A)^{r\wedge2}},
  \qquad
  A=c_r^{1/(r\wedge2)}Ld^{1+(1/q-1/2)_+-1/r}.
\]
Integrating this tail gives the Bernstein-type moment estimate: for every integer \(k\geq2\),
\[
  \sqrt{\E|\Delta(\zeta)|^{2k}}
  \leq
  \frac{k!}{2} C A^2 (C'A)^{k-2}\,,
\]
where \(C,C'\) depend only on \(r\).

It remains to understand the normal term. This part is purely geometric. We deal with $r=\infty$ separately, and do not detail it here; the corresponding normal has deterministic dual norm, making the proof straightforward. For $r \in [1,\infty)$,  $\zeta \in \partial B^d_r$, the normals admit the closed form expressions
\[
  n_1(\zeta)=\sign(\zeta),
  \qquad
  n_r(\zeta)_i=\sign(\zeta_i)|\zeta_i|^{r-1},
  \quad 1<r<\infty\,.
\]
The distribution of the coordinates under the cone measure is explicit. If $\zeta \sim \partial B^d_r$, then
\[
  |\zeta_i|^r\sim\betad(1/r,(d-1)/r)\,.
\]
This representation allows us to evaluate the normal moments sharply. There are two cases.

If \(p=1\), then \(p^*=\infty\), so the relevant quantity is the largest coordinate of the normal. For \(r=1\), the normal is a sign
vector and \(\norm{n_1(\zeta)}_\infty=1\). For \(r>1\), the beta tail bound and
a union bound show that \(\max_i|\zeta_i|^r\) is controlled at scale \((\log d)/d\). Consequently,
\[
  \sqrt{\E\norm{n_r(\zeta)}_\infty^4}
  \leq
  C_{r}d^{-2/r^*}(\log d)^{2/r^*}.
\]
The logarithm is the usual cost of taking a maximum over \(d\) coordinates, and
it is precisely the logarithm that later appears in the leading regret term when
\(p=1\).

If \(p>1\), the normal is measured in an averaged norm rather than a maximum
norm. Put \(T_i=|\zeta_i|^r\). For $r=1$, we again have a simple special case. For \(r>1\),
\[
  \norm{n_r(\zeta)}_{p^*}^{p^*}
  =
  \sum_{i=1}^d T_i^\beta,
  \qquad
  \beta=\frac{p^*}{r^*}.
\]
From here, the proof splits into cases depending on $\beta$ and $\alpha=4/p^*$, and uses convexity and concavity arguments together with estimates for the moments of beta random variables lead to
\[
  \sqrt{\E\norm{n_r(\zeta)}_{p^*}^4}
  \leq
  C_{p,r}d^{2(1/r-1/p)},
  \qquad p\in(1,\infty].
\]

Combining the bounds yields the Bernstein moment condition. For the formal proof, see \cref{appendix:grad-bounds}.

\subsection{On the peculiar case of $q > 2$: estimator suboptimality versus lower bound}\label{sec:q-suboptimal}

The upper bound of \cref{thm:upper-bound} and the lower bound of
\cref{thm:lower-bound} do not match when \(q>2\). Either our lower bound is loose, or randomised gradient estimators of the form we study are suboptimal in this regime. We suspect the latter. Indeed, consider the test function
\[
    f(x)=Lx_1 .
\]
This function is \(L\)-Lipschitz with respect to every \(\ell_q\) norm, since
its gradient is \(Le_1\), and $\norm{Le_1}_{q^*}=L$. For the two-point finite-difference magnitude, $ \Delta(\zeta) = Ld\zeta_1$. Thus even for this very simple linear function, the estimator amplifies the
single observed coordinate by the ambient dimension \(d\).

To see the consequence, take
\(r=p=2\). Then \(n_2(\zeta)=\zeta\) and
\(\norm{\zeta}_2=1\), so $g(\zeta)=Ld\zeta_1\zeta$. Consequently,
\[
 \nu_2 \geq \E\norm{g(\zeta)}_2^2
  =
  L^2d^2\E\zeta_1^2
  =
  L^2d .
\]
But this lower bound matches the $d$-dependence of the upper bound on $\nu_2$ for $q=2$; the estimator variance does not improve despite the additional restriction imposed by \(q>2\) Lipschitzness. This is in contrast with the lower bound of \cref{thm:lower-bound},
which is dimension-free when \(p=2\) and \(q=\infty\).

Our conjecture for why this happens is as follows. Examining the lower bound construction, we see that for $q > 2$, the hard instances are sparse, being supported on only \(s<d\) coordinates. This makes sense: \(\ell_\infty\)-Lipschitzness forces the \(\ell_1\) norm of every subgradient to be small, and so sparsity emerges. Yet, our finite difference gradient estimators have no mechanism for exploiting sparsity.

It remains an open question whether a different
zeroth-order estimator can attain the dimension dependence suggested by the
lower bound for \(q>2\), or whether the lower bound can be strengthened.

\printbibliography

\onecolumn
\appendix

\section{Gradient estimator dual norm moment bound}\label{appendix:grad-bounds}

For this appendix, fix $x \in \Rd$ and an $L$-Lipschitz function $f$ on $(\Rd, \norm{\cdot}_q)$. Write
\[
  \Delta(\zeta)
  =
  \frac{d}{2h}\roundb[\big]{f(x+h\zeta)-f(x-h\zeta)} \spaced{and} g(\zeta)=\Delta(\zeta)n_r(\zeta)\,.
\]
In this appendix we bound in turn the moments of $\Delta(\zeta)$ and $\norm{n_r(\zeta)}_{p^*}$, and then assemble the bound on the moments of $\norm{g(\zeta)}_{p^*}$, thus proving \cref{lem:gradient-moment-bound}. We deal the with $r=\infty$ case separately.

\subsection{Scalar gradient oscillation bound}

Recall the following concentration result for Lipschitz functions of cone measure random variables.

\lipschitzconcentration*

  We shall use this result to prove the following bound on the difference term $\Delta(\zeta)$.

\begin{lemma}\label{lem:diff-bound}
  Fix $r \in [1,\infty)$ and let $\alpha = r \wedge 2$. If $\zeta \sim \partial B^d_r$, then for every integer $k \geq 2$,
  \[
    \sqrt{\E |\Delta(\zeta)|^{2k}} \leq \frac{k!}{2} (12^{1/\alpha}\sqrt{C_r/6} A^2)  (2^{2/\alpha-1}A)^{k-2}\,,
  \]
  where
  \[
     A = c_r^{1/\alpha} L d^{1+(1/q-1/2)_+-1/r}\,.
  \]
\end{lemma}

\begin{proof}
  Let $\theta = (1/q-1/2)_+$. Observe that for any $\zeta, \zeta'$,
\begin{align*}
|\Delta(\zeta)-\Delta(\zeta')|
&\leq
\frac{d}{2h}
\left(
  |f(x+h\zeta)-f(x+h\zeta')|
  +
  |f(x-h\zeta')-f(x-h\zeta)|
\right)
\\
&\leq
dL\norm{\zeta-\zeta'}_q
\leq
d^{1+\theta}L\norm{\zeta-\zeta'}_2 ,
\end{align*}
  with the final inequality by norm equivalence on $\Rd$. Hence $\Delta$ is $d^{1+\theta}L$-Lipschitz on $(\Rd, \norm{\cdot}_2)$. Moreover, since $\Delta$ is odd and $\zeta$ symmetric, $\E \Delta(\zeta) = 0$. Therefore, by \cref{thm:lipschitz-concentration-1}, applied to the function $\phi(\zeta) = \Delta(\zeta) / (d^{1+\theta} L)$, we obtain that
  \[
    \P{|\Delta(\zeta)| > u} \leq C_r \exp\curlyb[\bigg]{-c_r^{-1} d^{\frac{2}{r\vee 2}} \roundb[\Big]{\frac{u}{d^{1+\theta}L}}^{r \wedge 2}} = C_r \exp\curlyb{-(u/A)^{\alpha}}\,.
  \]
  By the moment-tail identity for nonnegative random variables \citep[Lemma 2.4]{kallenberg1997foundations},
  \begin{align*}
        \E |\Delta(\zeta)|^{2k} = 2k \int_0^\infty u^{2k-1} \P{|\Delta(\zeta)| > u} \dif u &\leq  2k C_r \int_0^\infty u^{2k-1} e^{-(u/A)^{\alpha}} \dif u \\
        &= C_r A^{2k} \Gamma(1+2k/\alpha)\,,
  \end{align*}
  where the final equality follows by the definition of the gamma function. By the log-convexity of $\Gamma$,
  \[
    \Gamma(1+2k/\alpha) \leq \Gamma(1+k)^{2-2/\alpha} \Gamma(1+2k)^{2/\alpha - 1} = (k!)^{2-2/\alpha} ((2k)!)^{2/\alpha-1}\,.
  \]
  For $k \geq 2$, $k! \leq 2 \cdot \roundb{k!/2}^2$, and
  \[
    (2k)! = 24 \prod_{j=3}^k (2j-1)(2j) \leq 24\prod_{j=3}^k (2j)^2 = 24 \roundb{\prod_{j=3}2j}^2 = 24 \roundb{2^{k-2} \frac{k!}{2}}^{2}\,.
  \]
  Combining and simplifying the bounds on the $\Gamma$ term and taking the square root yields the result.
\end{proof}

\subsection{Moment bounds for the normal vector}

In this section, we prove the following moment bounds for the normal vector.

\begin{lemma}\label{lem:norm-moment-bound}
  Let $d \geq 2$, $r \in [1,\infty)$ and $\zeta \sim \partial B^d_r$. Then, for $p \in (1,\infty]$,
  \[
     \sqrt{\E[\norm{n_r(\zeta)}_{p^*}^4]} \leq  (1+r)^{1/2} (1 + (r-1)(p^* \vee 4))^{2/r^*}d^{2(1/r-1/p)}\,.
  \]
  Moreover,
  \[
    \sqrt{\E[\norm{n_r(\zeta)}_{\infty}^4]} \leq (10r \log d)^{2/r^*} d^{-2/r^*}\,.
  \]
\end{lemma}

We begin with the representation results used below, and then split into the
cases \(1\le p^*<\infty\) and \(p^*=\infty\). The following representation result was established independently by \citet{schechtman1990volume,rachev1991approximate}.

\begin{lemma}\label{lem:representation}
  Fix $r \in [1,\infty)$ and let $X_1, \dots,X_d$ be independent random variables with density $e^{-|t|^r}/(2\Gamma(1+1/r))$ for $t \in \R$. Consider the random vector $X = (X_1, \dots, X_d) \in \Rd$ and denote
  \[
    \zeta = \frac{X}{\norm{X}_r}\,.
  \]
  Then $\zeta$ is independent of $\norm{X}_r$. Moreover, $\zeta$ is distributed according to the cone measure on $\partial B^d_r$.
\end{lemma}

The proof of the following corollary is adapted from the proof of Theorem 3 in \citet{barthe2005probabilistic}.

\begin{corollary}\label{cor:representation-gamma}
  Let $X_1, \dots,X_d$ be independent random variables with density $e^{-|t|^r}/(2\Gamma(1+1/r))$ for $t \in \R$ (as in \cref{lem:representation}). Then $|X_i|^r \sim \gammad(1/r,1)$ for every $i=1,\dots,d$.
\end{corollary}

\begin{proof}
  Note that for each $i=1,\dots,d$, the density of $|X_i|^r$ is
  \[
    \frac{\dif}{\dif t} \P{|X_i| \leq t^{1/r}} = \frac{2t^{1/r-1}}{r} \cdot \frac{1}{2\Gamma(1+1/r)}e^{-t} = \frac{1}{\Gamma(1/r)} t^{1/r-1} e^{-t}\,, \qquad t > 0\,.
  \]
  Therefore, each $|X_i|^r$ has a $\gammad(1/r,1)$ distribution.
\end{proof}

We will need a number of standard facts relating gamma and beta random variables.

\begin{lemma}[\citet{ferguson1973bayesian}, p. 211]\label{lem:dist-facts}
  Let $Z_1,\dots, Z_k$ be independent random variables with $Z_i \sim \gammad(\alpha_i, 1)$ where $\alpha_i > 0$ for all $i$. Then,
  \[
    \frac{Z_i}{\sum_{i=1}^k Z_i} \sim \betad\roundb[\Big]{\alpha_i, \sum_{j \neq i} \alpha_j}\,, \qquad i=1,\dots,k\,,
  \]
  and
  \[
    \sum_{i=1}^k Z_i \sim \gammad\roundb[\Big]{\sum_{i=1}^k \alpha_i, 1}\,.
  \]
\end{lemma}

The above facts combined with the representation result give the following final representation.

\begin{proposition}
  Fix $r \in [1,\infty)$ and let $\zeta \sim \partial B^d_r$. Then $|\zeta_i|^r \sim \betad(1/r, \frac{d-1}{r})$ for every $i=1,\dots,d$.
\end{proposition}

\begin{proof}
By \cref{lem:representation}, write $\zeta = X/\norm{X}_r$, where each coordinate of $X=(X_1,\dots,X_d)$ is independent and has density $e^{-|t|^r}/(2\Gamma(1+1/r))$ for $t \in \R$. Then, by \cref{cor:representation-gamma}, each $|X_i|^r$ is an independent $\gammad(1/r,1)$ random variable, and therefore by \cref{lem:dist-facts},
\[
  |\zeta_i|^r = \frac{|X_i|^r}{\sum_{j=1}^d |X_j|^r}
\]
has a $\betad(1/r, \frac{d-1}{r})$ distribution for every $i=1,\dots,d$.
\end{proof}

We will also use these closed-form expressions for the normals of surface random variables.

\LemmaClosedFormNormal*

\begin{proof}[Proof of \cref{claim:normal}]
  Since $\zeta \in \partial B^d_r$, we have $\norm{\zeta}_r = 1$. By H\"older's inequality, for every $x \in B^d_{r^*}$,
  \[
    \langle x, \zeta\rangle \leq \norm{x}_{r^*}\norm{\zeta}_r \leq 1\,.
  \]

  Consider the case $r=1$. If $x = \sign(\zeta)$, then $\norm{x}_\infty = 1$, and
  \[
    \langle x, \zeta \rangle = \sum_{i=1}^d \sign(\zeta)_i \zeta_i = \sum_{i=1}^d |\zeta_i| = \norm{\zeta}_1 = 1\,.
  \]
  Therefore, $n_1(\zeta) = \sign(\zeta)$ is admissible as the normal for $\zeta \in \partial B^d_1$.

  Consider the case $1 < r < \infty$. If $x_i = \sign(\zeta_i)|\zeta_i|^{r-1}$, then
  \[
    \norm{x}_{r^\star}^{r^\star} = \sum_{i=1}^d |\zeta_i|^{(r-1)r^*} = \sum_{i=1}^d |\zeta_i|^r =\norm{\zeta}^r_r = 1\,,
  \]
  and
  \[
    \langle x, \zeta \rangle = \sum_{i=1}^d \sign(\zeta_i) \zeta_i |\zeta_i|^{r-1} = \sum_{i=1}^d |\zeta_i|^r = \norm{\zeta}_r^r = 1\,.
  \]
  Hence, taking $n_r(\zeta)$ defined elementwise by $n_r(\zeta)_i = \sign(\zeta_i)|\zeta_i|^{r-1}$ is admissible.

  Finally, consider the case \(r=\infty\). Since \(\zeta\in\partial B_\infty^d\),
  we have \(\norm{\zeta}_\infty=1\). By H\"older's inequality, for every
  \(x\in B_1^d\),
  \[
    \langle x,\zeta\rangle \leq \norm{x}_1\norm{\zeta}_\infty\leq1.
  \]
  If \(J\in\argmax_{i\le d}|\zeta_i|\), then
  \(x=\sign(\zeta_{J})e_{J}\) belongs to \(B_1^d\) and satisfies
  \[
    \langle x,\zeta\rangle=|\zeta_{J}|=\norm{\zeta}_\infty=1.
  \]
  Hence this choice is admissible as a normal. Under the cone measure on
  \(\partial B_\infty^d\), the maximising coordinate is unique almost surely, so
  this gives an unambiguous normal up to a null set.
\end{proof}

We are now ready to split into the two cases.

\subsubsection{Normal moment control, case $p^* = \infty$}

\begin{lemma}\label{lem:beta-concentration}
  If $X \sim \betad(a,b)$ with $a \in (0,1]$ and $b > 0$, then for all $u \in (0,1)$,
  \[
    \P{X \geq u} \leq (1-u)^{b}\,.
  \]
\end{lemma}

\begin{proof}
 If $Y \sim \betad(1,b)$, then for every $u \in (0,1)$,
  \[
    \P{Y \geq u} = \frac{1}{B(1,b)}\int_u^1(1-t)^{b-1} \dif t = (1-u)^b\,.
  \]
This covers the case $a=1$. Now suppose $a \in (0,1)$. Let $Z_1 \sim \gammad(a,1)$, $Z_2 \sim \gammad(1-a,1)$, $N\sim\gammad(b,1)$ be independent random variables. By \cref{lem:dist-facts}, we may realise
  \[
    X = \frac{Z_1}{Z_1+N} \sim \betad(a,b)\,, \qquad Y = \frac{Z_1+Z_2}{Z_1+ Z_2+N} \sim \betad(1,b)\,.
  \]
  Since for any $x,y \geq 0$ with $x+y > 0$, $\delta \mapsto \frac{x+\delta}{x+y+\delta}$ is an increasing function on $[0,\infty)$, $X \leq Y$ almost surely, and therefore $\P{X \geq u} \leq \P{Y \geq  u}$ for all $u \in (0,1)$.
\end{proof}

\begin{proof}[Proof of \cref{lem:norm-moment-bound}, case $p^*=\infty$]
  For $r=1$, $r^*=\infty$, and $n_1(\zeta) = \sign(\zeta)$, so $\norm{n_r(\zeta)}_{\infty} = 1$.

  Consider $r > 1$. Let $M = \max_{i=1,\dots,d} |\zeta_i|^r$ and $\beta = \frac{4}{r^*}\in (0,4)$. By \cref{claim:normal}, $n_r(\zeta)_i = \sign(\zeta_i)|\zeta_i|^{r-1}$, we seek to bound
  \[
    \sqrt{\E\norm{n_r(\zeta)}_\infty^4} = \roundb{\E \max_{i \leq d} |\zeta_i|^{4(r-1)}}^{\frac{1}{2}} = \norm{M}_{L_\beta}^{\beta/2}\,.
  \]
  Since for $i=1,\dots,d$, $|\zeta_i|^r \sim \betad(1/r,b)$ with $b=\frac{d-1}{r}$, by \cref{lem:beta-concentration} and the union bound, for $0 < u < b$,
  \[
    \P{bM \geq u} \leq 1\wedge \sum_{i=1}^d\P{|\zeta_i|^r \geq u/b} \leq 1 \wedge d(1-u/b)^b \leq 1 \wedge de^{-u}\,.
  \]
  And since $M \leq 1$, for $u \geq b$, $\P{bM \geq u} = 0$. On the other hand, if  $E \sim \operatorname{Exp}(1)$, then
  \[
    \P{E+\log d \geq u} = 1 \wedge e^{-(u-\log d)} = 1 \wedge de^{-u}\,.
  \]
  Thus, we have the stochastic domination $\P{bM \geq u} \leq \P{E+\log d \geq u}$, and hence for every $\beta > 0$,
  \[
    b^\beta \E[M^\beta] \leq \E[(E+\log d)^\beta]\,.
  \]
  If $\beta \in (0,1]$, then $x \mapsto x^\beta$ is concave, so Jensen's gives
  \[
    \E(E+\log d)^\beta \leq (\E[E] + \log d)^\beta = (\log d+1)^\beta\,.
  \]
  If $\beta \in [1,4]$, by Minkowski's inequality and monotonicity of \(L_\beta\) norms on probability spaces,
  \[
    \norm{\log d+E}_{L_\beta} \leq \log d + \norm{E}_{L_\beta} \leq \log d + \norm{E}_{L_4} = \log d+\Gamma(5)^{1/4} = \log d+24^{1/4}\,,
  \]
  where at the end we used the standard expression for moments of exponential random variables. Putting the inequalities together, we obtain that $\norm{M}_{L_\beta} \leq b^{-1}(\log d + 24^{1/4})$. The reported bound follows by using $d \geq 2$ to relax this result.
\end{proof}

\subsubsection{Normal moment control, case $p^* \in [1,\infty)$}

\begin{lemma}\label{lem:gamma-ratio-bound}
  Let $x>0$ and $u \geq 0$. Then
  \[
    \frac{x^u}{1+1/x} \leq \frac{\Gamma(x+u)}{\Gamma(x)} \leq (x+u)^u\,.
  \]
\end{lemma}

\begin{proof}
  Write $u = m+\theta$, $m \in \N$, $\theta \in [0,1)$. Then
  \[
    \frac{\Gamma(x+u)}{\Gamma(x)} = \frac{\Gamma(x+\theta)}{\Gamma(x)} \prod_{j=0}^{m-1} (x+\theta+j)\,.
  \]
  By Gautschi's inequality (DLMF \S5.6.4) and using that $\Gamma(x+1)=x\Gamma(x)$, for $0 < \theta < 1$,
  \[
    \frac{x}{(x+1)^{1-\theta}} < \frac{\Gamma(x+\theta)}{\Gamma(x)} < x^\theta\,,
  \]
  where
  \[
     \frac{x}{(x+1)^{1-\theta}} =  x^\theta \roundb[\bigg]{\frac{x}{x+1}}^{1-\theta} \geq \frac{x^\theta}{1+1/x}\,.
  \]
  If $\theta = 0$, same bounds apply with weak inequality in place of strict. Therefore,
  \[
    \frac{x^{u}}{1+1/x} = \frac{x^\theta}{1+1/x} x^{m} \leq \frac{\Gamma(x+\theta)}{\Gamma(x)} \prod_{j=0}^{m-1} (x+\theta+j) \leq x^\theta (x+u)^m \leq (x+u)^{u},
  \]
  which yields the result after identifying the middle term with $\Gamma(x+u)/\Gamma(x)$.
\end{proof}

\begin{proof}[Proof of \cref{lem:norm-moment-bound}, case $p^* \in [1,\infty)$]
  If $r = 1$, then $r^* = \infty$ and $n_1(\zeta) = \sign(\zeta)$. Therefore, $\norm{n_1(\zeta)}_{p^*} = d^{1/p^*}$, and
  \[
    (\E \norm{n_1(\zeta)}^{4}_{p^*})^{1/2} = d^{2/p^*} = d^{2(1/r-1/p)}\,.
  \]

  Now suppose $r > 1$. Let $a = 1/r$, $b=(d-1)/r$ and $T_i = |\zeta_i|^r$. Note $T_i \sim \betad(a,b)$ for $i=1,\dots,d$. Write $\beta = \frac{(r-1)p^*}{r} > 0$. From the closed-form for the normal, \cref{claim:normal},
  \[
    N(\zeta) := \norm{n_r(\zeta)}_{p^*}^{p^*} = \sum_{i=1}^d |\zeta_i|^{(r-1)p^*} = \sum_{i=1}^d T_i^\beta\,.
  \]
  Letting $\alpha = 4/p^*$ we look to bound \[
    \sqrt{\E \norm{n_r(\zeta)}_{p^*}^4} = \roundb{\E N(\zeta)^\alpha}^{1/2}\,.
  \]

  Suppose first that $\beta \in (0,1]$. Then $x \mapsto x^\beta$ is concave, and since $\sum_{i=1}^d T_i = 1$,
  \[
    N(\zeta) = \sum_{i=1}^d T_i^\beta \leq d\roundb[\bigg]{\frac{1}{d} \sum_{i=1}^d T_i}^\beta = d^{1-\beta}\,.
  \]
  Therefore,
  \[
    \roundb{\E N(\zeta)^\alpha}^{1/2} \leq d^{\alpha(1-\beta)/2} = d^{2(1/r-1/p)}\,.
  \]

  Now suppose that $\beta > 1$. There are two subcases. First assume that $p^* \leq 4$, such that $\alpha \geq 1$. By Jensen's inequality applied to the function $x \mapsto x^\alpha$,
  \[
    N(\zeta)^\alpha = \roundb[\Big]{\sum_{i=1}^d T_i^\beta}^\alpha = d^\alpha \roundb[\Big]{\frac{1}{d}\sum_{i=1}^d T_i^\beta}^\alpha \leq d^{\alpha-1}\sum_{i=1}^d T_i^{\beta \alpha}\,.
  \]
  Taking expectations and using exchangeability,
  \[
    \E N(\zeta)^\alpha \leq d^\alpha \E T_1^{\beta \alpha} = d^{\alpha} \E T_1^{4/r^*}\,.
  \]
  Write $s := \beta \alpha = 4/r^*$. Since $T_1 \sim \betad(a,b)$,
  \[
    \E T_1^{s} = \frac{1}{B(a,b)} \int_0^1 t^{a+s-1}(1-t)^{b-1} \dif t = \frac{B(a+s,b)}{B(a,b)}\,,
  \]
  where $B$ denotes the Beta integral function. Using the identity $B(x,y) = \frac{\Gamma(x)\Gamma(y)}{\Gamma(x+y)}$, this yields
  \[
    \E N(\zeta)^\alpha \leq d^\alpha \E T_1^s \leq d^\alpha \cdot \frac{\Gamma(da)}{\Gamma(da + s)} \cdot \frac{\Gamma(a + s)}{\Gamma(a)}\,.
  \]
  Applying the two inequalities of \cref{lem:gamma-ratio-bound} to the two ratios yields
  \begin{align*}
    \E N(\zeta)^\alpha
    &\leq d^\alpha (1+1/(da))(da)^{-s} (a+s)^{s} \\
    &= d^{\alpha(1-\beta)} (1+r/d)\roundb{1+4(r-1)}^{s} \\
    &\leq d^{\alpha(1-\beta)}(1+r)(1+4(r-1))^{s}\,.
  \end{align*}

  For the second subcase, assume that $p^* > 4$, such that $\alpha < 1$. Then $x \mapsto x^\alpha$ is concave, and
  \[
    \E N(\zeta)^\alpha \leq (\E N(\zeta))^\alpha\,.
  \]
  Using exchangeability and the same beta moment estimate as above, with $\beta$ in place of $s$, we obtain
  \[
    (\E N(\zeta))^\alpha = (d\E T_1^\beta)^\alpha \leq d^{\alpha(1-\beta)} (1+r/d)^\alpha (1+r\beta)^{\beta \alpha}
  \]
  Taking the square root gives the bound for this case. Combining the cases yields the final bound.
\end{proof}

\subsection{The cube endpoint \(r=\infty\)}\label{appendix:cube-endpoint}

The endpoint \(r=\infty\) is not covered by the beta/gamma representation used
above. It is nevertheless elementary. The cone measure on the cube is described
as follows.

\begin{lemma}[Cone measure on the cube]\label{lem:cube-cone-representation}
Let \(J\) be uniform on \(\curlyb{1,\dots,d}\), let \(S\) be uniform on
\(\curlyb{-1,1}\), and let \(U_1,\dots,U_d\) be independent
\(\operatorname{Unif}[-1,1]\) random variables, independent of \(J,S\). Define
\(\zeta\in\Rd\) by
\[
  \zeta_J=S \spaced{and}
  \zeta_i=U_i \quad \text{for } i\neq J.
\]
Then \(\zeta\) is distributed according to the cone measure on
\(\partial B_\infty^d\). Moreover, \(J\) is the unique coordinate satisfying
\(|\zeta_J|=1\) almost surely, and
\[
  n_\infty(\zeta)=S e_J .
\]
\end{lemma}

\begin{proof}
For \(j\le d\) and \(s\in\curlyb{-1,1}\), let $F_{j,s}=\curlyb{z\in\partial B_\infty^d: z_j=s}$ be a face of the cube. If \(A\subset F_{j,s}\) is measurable, then the cone over
\(A\) has volume
\[
  \vol\curlyb{ta:a\in A,0\leq t\leq1}
  =
  \int_A\int_0^1 t^{d-1}\dif t\,\dif a
  =
  \frac{1}{d}\vol_{d-1}(A).
\]
Since the cube has volume \(2^d\), the cone measure assigns mass
\(\vol_{d-1}(A)/(d2^d)\) to \(A\). In particular, the face is chosen uniformly
among the \(2d\) faces, and conditional on the face, the remaining coordinates
are uniform on \([-1,1]\). This proves the representation. The expression for
the normal follows from the \(r=\infty\) case of \cref{claim:normal}.
\end{proof}

The moment bound at $r=\infty$ is simpler than for finite \(r\), because
the normal is coordinate-valued.

\begin{lemma}[Endpoint scalar moment bound]\label{lem:cube-endpoint-moment}
Let \(d\ge2\), \(q\in[1,\infty]\), and let
\(\zeta\sim\partial B_\infty^d\) be distributed according to the cone measure.
Put
\[
  \theta=(1/q-1/2)_+,
  \qquad
  \gamma_\infty=(q\vee2)^{-1},
  \qquad
  \sqrt{\nu_\infty}=\sqrt{2}Ld^{1+\theta}.
\]
Then, for every integer \(k\ge2\),
\[
  \E|\Delta(\zeta)|^k
  \leq
  \frac{k!}{2}\nu_\infty\roundb{d^{\gamma_\infty}\sqrt{\nu_\infty}}^{k-2}.
\]
\end{lemma}

\begin{proof}
Let $L_0=Ld^{1+\theta}$. As in the finite-\(r\) case, for any \(\zeta,\zeta'\in\partial B_\infty^d\),
\[
  |\Delta(\zeta)-\Delta(\zeta')|
  \leq
  dL\norm{\zeta-\zeta'}_q
  \leq
  L_0\norm{\zeta-\zeta'}_2.
\]
Hence \(\Delta\) is \(L_0\)-Lipschitz on each face of the cube, with respect to
the Euclidean metric on that face.

Condition on \(J=j\) and \(S=s\), using the notation of
\cref{lem:cube-cone-representation}. On this face, the free coordinates are
distributed according to the product uniform measure on \([-1,1]^{d-1}\).
Therefore, by the Poincaré inequality for the cube and tensorisation
\citep[see, e.g.,][Chapter 5]{ledoux2001concentration},
\[
  \var(\Delta(\zeta)\mid J=j,S=s)\leq L_0^2.
\]
Let
\[
  m_{j,s}=\E[\Delta(\zeta)\mid J=j,S=s].
\]
Flipping only the \(j\)-th coordinate maps the face \((j,s)\) to the face
\((j,-s)\) and moves every point by Euclidean distance \(2\). Since \(\Delta\)
is \(L_0\)-Lipschitz,
\[
  |m_{j,s}-m_{j,-s}|\leq 2L_0.
\]
On the other hand, by the change of variables \(U\mapsto -U\) on the opposite
face, central symmetry of the cone measure and oddness of \(\Delta\) give
\(m_{j,-s}=-m_{j,s}\). Hence \(|m_{j,s}|\leq L_0\). Combining the
conditional variance and conditional mean bounds gives
\[
  \E\Delta(\zeta)^2
  =
  \E\var(\Delta(\zeta)\mid J,S)+\E m_{J,S}^2
  \leq
  2L_0^2
  =
  \nu_\infty.
\]

It remains to pass from second moments to all moments. Since
\(\norm{\zeta}_\infty=1\), we have
\[
  |\Delta(\zeta)|
  \leq
  dL\norm{\zeta}_q
  \leq
  Ld^{1+1/q}
  =
  Ld^{1+\theta+{\gamma_\infty}}
  \leq
  d^{\gamma_\infty}\sqrt{\nu_\infty}.
\]
Therefore, for every integer \(k\ge2\),
\[
  \E|\Delta(\zeta)|^k
  \leq
  \roundb{d^{\gamma_\infty}\sqrt{\nu_\infty}}^{k-2}\E\Delta(\zeta)^2
  \leq
  \nu_\infty\roundb{d^{\gamma_\infty}\sqrt{\nu_\infty}}^{k-2}
  \leq
  \frac{k!}{2}\nu_\infty\roundb{d^{\gamma_\infty}\sqrt{\nu_\infty}}^{k-2}.
\]
\end{proof}

We also verify that the $\nu_\infty$ estimate is tight, as follows.

\begin{proposition}[Tightness of the cube-endpoint variance proxy]\label{prop:cube-endpoint-tight}
Let \(d\ge2\), \(p,q\in[1,\infty]\), and let
\(\zeta\sim\partial B_\infty^d\) be distributed according to the cone measure on
the cube. Put
\[
  \theta=(1/q-1/2)_+,
  \qquad
  \gamma_\infty=(q\vee2)^{-1},
  \qquad
  \sqrt{\nu_\infty}=\sqrt{2}Ld^{1+\theta}.
\]
For \(a\in\Rd\), let \(f_a(x)=\langle a,x\rangle\), and let \(g_a(\zeta)\) be
the corresponding cube-endpoint two-point estimator. Then
\[
  \sup_{\norm{a}_{q^*}\le L}\E\norm{g_a(\zeta)}_{p^*}^2
  =
  \frac{d(d+2)}{3}L^2d^{2\theta}.
\]
In particular, there exists an \(L\)-Lipschitz linear function \(f_a\) such that
\[
  \E\norm{g_a(\zeta)}_{p^*}^2
  \geq
  \frac{1}{6}\nu_\infty.
\]
Consequently, the variance proxy \(\nu_\infty\) for the cube endpoint is sharp
up to a universal numerical constant.
\end{proposition}

\begin{proof}
By the cube-cone representation, we may write \(\zeta\) by first choosing
\(J\) uniformly from \(\curlyb{1,\dots,d}\), then choosing a Rademacher sign
\(S\), setting \(\zeta_J=S\), and sampling the remaining coordinates uniformly
from \([-1,1]\). The endpoint normal is \(n_\infty(\zeta)=Se_J\), and hence
\(\norm{n_\infty(\zeta)}_{p^*}=1\) for every \(p\in[1,\infty]\). For the linear function \(f_a(x)=\langle a,x\rangle\), the finite-difference
magnitude is
\[
  \Delta(\zeta)
  =
  \frac{d}{2h}\roundb{f_a(x+h\zeta)-f_a(x-h\zeta)}
  =
  d\langle a,\zeta\rangle,
  \qquad
  \norm{g_a(\zeta)}_{p^*}
  =
  d|\langle a,\zeta\rangle|.
\]
Moreover, the cube-cone coordinates satisfy
\[
  \E\zeta_i^2=\frac1d+\frac{d-1}{d}\cdot\frac13=\frac{d+2}{3d},
  \qquad
  \E\zeta_i\zeta_j=0 \quad \text{for } i\neq j.
\]
Therefore,
\[
  \E\norm{g_a(\zeta)}_{p^*}^2
  =
  d^2\E\langle a,\zeta\rangle^2
  =
  d^2\sum_{i=1}^d a_i^2\E\zeta_i^2
  =
  \frac{d(d+2)}{3}\norm{a}_2^2.
\]

It remains to maximise \(\norm{a}_2\) subject to \(\norm{a}_{q^*}\le L\). By
norm equivalence,
\[
  \sup_{\norm{a}_{q^*}\le L}\norm{a}_2
  =
  Ld^{(1/2-1/q^*)_+}
  =
  Ld^{(1/q-1/2)_+}
  =
  Ld^\theta.
\]
This supremum is attained, for example, by taking \(a=Le_1\) when \(q\ge2\),
and by taking \(a=Ld^{-1/q^*}\one\) when \(q<2\),  where $\one \in \Rd$ denotes the all ones vector. Hence,
\[
  \sup_{\norm{a}_{q^*}\le L}\E\norm{g_a(\zeta)}_{p^*}^2
  =
  \frac{d(d+2)}{3}L^2d^{2\theta}.
\]
Since \(d\ge2\),
\[
  \frac{d(d+2)}{3}L^2d^{2\theta}
  \geq
  \frac13 L^2d^{2+2\theta}
  =
  \frac16\nu_\infty,
\]
which proves the claimed lower bound. Finally, \(f_a\) is \(L\)-Lipschitz with
respect to \(\norm{\cdot}_q\) whenever \(\norm{a}_{q^*}\le L\), so the example
belongs to the problem class.
\end{proof}

\subsection{Gradient dual norm moment bound}

Recall our choice of proxy variance, and the statement of the dual moment bound proposition.

\defVar*

\propGradBound*

\begin{proof}
  Fix \(t\ge1\), and condition on \(\FF_{t-1}\vee\sigma(f_t)\). Then
  \(x:=x_t\), \(h:=h_t\), and \(f:=f_t\) are fixed, and the only randomness is
  \(\zeta:=\zeta_t\sim\partial B_r^d\). Since the resulting upper bound is deterministic,
taking conditional expectation again with respect to \(\FF_{t-1}\) gives the
required \(\SBM\) bound.

  Consider the case $r=\infty$. By
  \cref{lem:cube-cone-representation}, the normal is coordinate-valued, and so
  \[
    \norm{n_\infty(\zeta)}_{p^*}=1
    \spaced{and}
    \norm{g(\zeta)}_{p^*}=|\Delta(\zeta)|.
  \]
  Applying \cref{lem:cube-endpoint-moment} gives, for every \(k\ge2\),
  \[
    \E\squareb{\norm{g(\zeta)}_{p^*}^k}
    =
    \E|\Delta(\zeta)|^k
    \leq
    \frac{k!}{2}\nu_\infty\roundb{d^{\gamma_\infty}\sqrt{\nu_\infty}}^{k-2}\,.
  \]
  Therefore, with $r=\infty$, the process satisfies \(\operatorname{SBM}(\nu_\infty,d^{\gamma_\infty}\sqrt{\nu_\infty})\).

  Now suppose $r \in [1,\infty)$. Let \(\alpha = r \wedge 2\). Write $\theta = (1/q-1/2)_+$. We may assume \(C_r \geq 1\), since enlarging
  \(C_r\) preserves \cref{thm:lipschitz-concentration-1}. Write
  \[
    b_{p,r}
    =
    \begin{cases}
      (10r)^{1/r^*}, & p=1,\\
      (1+r)^{1/4}\roundb{1+(r-1)(p^*\vee4)}^{1/r^*},
      & p \in (1,\infty]\,.
    \end{cases}
  \]
  Put
  \[
    A=c_r^{1/\alpha}L d^{1+\theta-1/r}.
  \]
  By \cref{lem:diff-bound}, for every \(k\geq2\),
  \[
    \sqrt{\E |\Delta(\zeta)|^{2k}}
    \leq
    \frac{k!}{2}
    \roundb[\big]{12^{1/\alpha}\sqrt{C_r/6}\,A^2}
    \roundb[\big]{2^{2/\alpha-1}A}^{k-2}.
  \]

  Write $N(\zeta) := \norm{n_r(\zeta)}_{p^*}$. By
  \cref{lem:norm-moment-bound},
  \[
    \roundb{\E N(\zeta)^4}^{1/4}
    \leq
    b_{p,r}D_{p,r}(d)d^{1/r-1/p}.
  \]
  And since $n_r(\zeta) \in B^d_{r^*}$, by norm equivalence,
  \[
    N(\zeta) = \norm{n_r(\zeta)}_{p^*} \leq d^{(1/r-1/p)_+} \norm{n_r(\zeta)}_{r^*} = d^{(1/r-1/p)_+}\,.
  \]
  Therefore, for every \(k\geq2\),
  \[
    \sqrt{\E N(\zeta)^{2k}}
    \leq \squareb[\big]{d^{(1/r-1/p)_+}}^{(k-2)}
    b_{p,r}^2D_{p,r}(d)^2d^{2(1/r-1/p)}.
  \]

  Applying Cauchy-Schwarz gives
  \begin{align*}
    \E&\norm{g(\zeta)}_{p^*}^k
    =
    \E\squareb[\big]{|\Delta(\zeta)|^kN(\zeta)^k}
    \leq
    \sqrt{\E |\Delta(\zeta)|^{2k}}
    \sqrt{\E N(\zeta)^{2k}}
    \\
    &\leq
    \frac{k!}{2}
    \roundb[\big]{12^{1/\alpha}\sqrt{C_r/6}}
    \roundb[\big]{
      c_r^{1/\alpha}Lb_{p,r}D_{p,r}(d)d^{1+\theta-1/p}
    }^2
    \cdot
    \roundb[\big]{
      2^{2/\alpha-1}
      c_r^{1/\alpha}Ld^{1+\theta-1/r+(1/r-1/p)_+}
    }^{k-2}.
  \end{align*}

  Since \(\alpha\in[1,2]\),
  \[
    \roundb[\big]{12^{1/\alpha}\sqrt{C_r/6}}^{1/2}
    =
    12^{1/(2\alpha)}6^{-1/4}C_r^{1/4}
    \leq
    4C_r^{1/4}.
  \]
  Hence, by the definition of \(\nu_r\),
  \[
    \roundb[\big]{12^{1/\alpha}\sqrt{C_r/6}}
    \roundb[\big]{
      c_r^{1/\alpha}Lb_{p,r}D_{p,r}(d)d^{1+\theta-1/p}
    }^2
    \leq
    \nu_r.
  \]

  It remains to compare the Bernstein scale. Since
  \[
    1+\theta-1/r+\roundb[\big]{1/r-1/p}_+
    =
    1+\theta-1/p+\roundb[\big]{1/p-1/r}_+
    =
    1/p^*+\theta+{\gamma_r},
  \]
  and since \(2^{2/\alpha-1}\leq2\), while
  \(b_{p,r}D_{p,r}(d)\geq1\) for \(d\geq2\), we have
  \[
    2^{2/\alpha-1}
    c_r^{1/\alpha}Ld^{1+\theta-1/r+(1/r-1/p)_+}
    \leq
    d^{\gamma_r}\sqrt{\nu_r}.
  \]
  Combining the preceding displays yields
  \[
    \E\squareb[\big]{\norm{g(\zeta)}_{p^*}^k}
    \leq
    \frac{k!}{2}\nu_r\roundb[\big]{d^{\gamma_r}\sqrt{\nu_r}}^{k-2}.
  \]
  Since \(t\geq1\) and the integer \(k\geq2\) were arbitrary, the process
  \(\norm{g_1}_{p^*},\norm{g_2}_{p^*},\dots\) satisfies
  \(\operatorname{SBM}(\nu_r,d^{\gamma_r}\sqrt{\nu_r})\).
\end{proof}
\clearpage

\section{Concentration results}
\label{appendix:concentration}

  \begin{definition}\label{def:subgamma}
    Let $\nu, c>0$ and let $\bA$ be a $\sigma$-field. A random variable $X$ is $\bA$-conditionally $(\nu, c)$-subgamma if
    \[
      \log \E[e^{sX} \mid \bA] \leq \frac{\nu s^2}{2(1-c|s|)} \quad \text{for all $s \in (-1/c,1/c)$}\,.
    \]
  \end{definition}
  To establish said tail condition, we will use Bernstein's moment condition.
  \begin{lemma}[Bernstein's moment condition]\label{lem:bernstein}
  Let $\bA$ be a $\sigma$-field. Fix $\nu,c > 0$ and let $X$ be a random variable. Suppose that for every integer $k \geq 2$
  \[
    \E \big[|X|^k \mid \bA \big] \leq \frac{k!}{2} \nu c^{k-2}\,.
  \]
  Then $X - \E[X\mid\bA]$ is $\bA$-conditionally $(\nu,c)$-subgamma.
  \end{lemma}

\begin{proof}
  This is a special case of Theorem 2.10 in \citet{boucheron-concentration}.
\end{proof}

\subsection{Subgamma process concentration}

\begin{theorem}[Theorem 3.1, \citet{whitehouse2023time}]
    \label{thm:subpsi}
Fix $\delta \in (0,1)$ and $\nu,c > 0$. Let $(\bA_t)$ be a filtration and $(X_t)$ be a random process such that for all $t$, $X_t$ is $\bA_{t-1}$-conditionally $(\nu,c)$-subgamma. Let $\alpha > 1$, $\beta > 0$, and let $h: \Rp \rightarrow \Rp$ be an increasing function such that $\sum_{k \in \N} 1/h(k) \leq 1$. Define the function $\ell_\beta: \Rp \rightarrow \Rp$ by
    \[
        \ell_{\beta}(r)  = \log h \left(\log_{\alpha} \left( \frac{r \vee \beta}{\beta} \right) \right) + \log\left(\frac{2}{\delta}\right),
    \]
    where, for brevity, we have suppressed the dependence of $\ell_{\beta}$ on $\alpha$ and $h$. Let $\psi(\lambda) = \frac{\lambda^2}{2(1-\lambda |c|)}$. Then,
    \[
        \P[\bigg]{\exists n \geq 1 \colon |\sum_{t=1}^n X_t| \geq \left( n \nu \vee \beta
            \right) \cdot (\psi^*)^{-1} \roundb[\bigg]{\frac{\alpha\ell_{\beta}(n \nu)}{n\nu \vee \beta}
            }}\leq \delta\,,
    \]
    where $\psi^*$ denotes the convex conjugate of $\psi$.
\end{theorem}

\subgammaconcentration*

\begin{proof}[Proof of \cref{thm:bernstein-process-concentration}]
    Write $\Lambda_n := \Lambda_n(\delta)$. Apply \cref{thm:subpsi} with $\alpha = e$, $\beta = e^2c^2$ and $h(r) = (r+2)^2$. Write $x_n = \frac{n\nu}{c^2}$, where we note that $\sum_{k \in \N} 1/h(k) = \sum_{k=3}^\infty \frac{1}{k^2} < 1$. Then,
    \[
      \ell_\beta(n\nu) = 2\log\roundb[\Big]{2+\log\roundb[\Big]{\frac{x_n \vee e^2}{e^2}}} + \log(2/\delta)\,.
    \]
    If $x_n \leq e^2$, then
    \[
      \ell_\beta(n\nu) = 2\log2 + \log(2/\delta) \leq \Lambda_n\,.
    \]
    Likewise, if $x_n > e^2$, then
    \[
      2+\log\roundb[\Big]{\frac{x_n \vee e^2}{e^2}} = \log x_n\,,
    \]
    and therefore
    \[
      \ell_\beta(n\nu) = 2\log\log x_n + \log 2/\delta \leq \Lambda_n\,.
    \]
    That is, for all $n$, $\ell_\beta(n\nu) \leq \Lambda_n$.

    Now recall that $(\psi^*)^{-1}(u) = \sqrt{2u} + cu$. Therefore, on the complement event of \cref{thm:subpsi}, for all $n \geq 1$,
    \[
      \abs{\sum_{t=1}^n X_t} < \sqrt{2e(n\nu \vee e^2c^2) \ell_\beta(n\nu)} + ec\ell_\beta(n\nu)\,.
    \]
    Using that $n\nu \vee e^2c^2 \leq n\nu + e^2c^2$, that $\ell_\beta(n\nu) \leq \Lambda_n$, and that since $\Lambda_n > 1$, we have that $\sqrt{\Lambda_n} \leq \Lambda_n$, the right-hand side can be upper-bounded by
    \[
      \sqrt{2e}\sqrt{n\nu\Lambda_n} + (e\sqrt{2e} + e) c\Lambda_n\,.
    \]
    Finally, use $\sqrt{2e} < 3$ and $e\sqrt{2e} + e < 10$.
\end{proof}

\subsection{Concentration for processes with squared subgamma increments}

\weibullTheorem*

\begin{proof}[Proof of \cref{thm:weibul-time-uniform}]
    Write $\bP_{t-1}$ and $\E_{t-1}$ for conditional probability and conditional expectation given $\bA_{t-1}$. For $n \geq 1$, define
  \[
    A_n = 2 \log\roundb{4n} + \log\roundb{1/\delta}\,, \qquad
    B_n = \sqrt{\nu} + \sqrt{2\nu A_n} + cA_n\,.
  \]
  From the moment assumptions,
    \[
      \E_{t-1} X_t^2 \leq \nu\,, \qquad \E_{t-1} X_t \leq \sqrt{\E_{t-1} X_t^2} \leq \sqrt{\nu}\,.
    \]
    From the Bernstein to subgamma lemma, \cref{lem:bernstein}, the centred variable $X_t - \E_{t-1} X_t$ is conditionally $(\nu,c)$-subgamma. Hence, by the standard Chernoff bound for subgamma random variables applied conditionally \citep[see, for example,][Section 2.4]{boucheron-concentration}, for every $r \geq 0$,
    \[
      \bP_{t-1}\curlyb{X_t - \E_{t-1} X_t > \sqrt{2\nu r} + cr}
      \leq e^{-r}\,.
    \]
    Using the bound on $\E_{t-1} X_t \leq \sqrt{\nu}$ and taking expectations, since $B_t = \sqrt{\nu} + \sqrt{2\nu A_t} + cA_t$, we obtain that
    \[
      \P{X_t >B_t} \leq e^{-A_t} = \frac{\delta}{16t^2}\,.
    \]
    Therefore,
    \[
      \P{\exists t \geq 1 \colon X_t > B_t} \leq \sum_{t=1}^\infty \frac{\delta}{16t^2} = \frac{\pi^2}{96}\delta\,.
    \]

    Now fix $n \geq 1$ and define $Z_{t,n} = X_t^2 \wedge B_n^2$ for $1 \leq t \leq n$, and let
    \[
      M_{k,n} = \sum_{t=1}^k \roundb{Z_{t,n} - \E_{t-1}Z_{t,n}}\,,
      \qquad 0 \leq k \leq n\,.
    \]
    Then $(M_{k,n})_{0 \leq k \leq n}$ is a martingale. Its increments satisfy
    \[
      \Delta_{t,n} = Z_{t,n} - \E_{t-1}Z_{t,n} \leq Z_{t,n} \leq B^2_n\,.
    \]
    Moreover,
    \[
      \Var_{t-1}\roundb{\Delta_{t,n}} \leq
      \E_{t-1} Z_{t,n}^2 \leq B_n^2 \E_{t-1} Z_{t,n} \leq B_n^2 \E_{t-1} X_t^2 \leq B_n^2 \nu.
    \]
    Thus,
    \[
      \sum_{t=1}^n \Var_{t-1}\roundb{\Delta_{t,n}}
      \leq
      n B_n^2 \nu .
    \]
    By Freedman's inequality \citep[Theorem 1.6]{freedman1975tail},
    \[
      \P{M_{n,n} > B_n\sqrt{2\nu n A_n} + B_n^2A_n/3} \leq e^{-A_n}\,.
    \]
    Taking the union bound over $n$ gives
    \[
      \P{\exists n \geq 1 \colon M_{n,n} > B_n\sqrt{2\nu n A_n} + B_n^2A_n/3 } \leq \sum_{n=1}^\infty e^{-A_n} = \frac{\pi^2}{96}\delta\,.
    \]

    Now work on the event where for all $ n \geq 1$,
    \[
      X_n \leq B_n \spaced{and} M_{n,n} \leq B_n\sqrt{2\nu n A_n} + B_n^2A_n/3\,,
    \]
    the probability of which is at least $1-\frac{\pi^2}{48} \delta \geq 1-\delta$. Fix $n \geq 1$. Since $A_t$ and $B_t$ are increasing in $t$, for every $t \leq n$,
    \[
      X_t \leq B_t \leq B_n\,.
    \]
    Hence, $Z_{t,n} = X_t^2$ for all $t \leq n$. Therefore,
    \[
      \sum_{t=1}^n X_t^2 = \sum_{t=1}^n Z_{t,n} = \sum_{t=1}^n \E_{t-1}[Z_{t,n}] + M_{n,n} \leq \sum_{t=1}^n \E_{t-1} X_t^2 + B_n\sqrt{2\nu n A_n} + \frac{B^2_n}{3}A_n\,.
    \]
        Since $\sum_{t=1}^n \E_{t-1} X_t^2 \leq n\nu$, we have
    \[
      \sum_{t=1}^n X_t^2
      \leq
      n\nu + B_n\sqrt{2\nu n A_n} + \frac{B_n^2}{3}A_n .
    \]
    It remains to take square roots. Set
    \[
      a_n := \sqrt{\nu n},
      \qquad
      D_n := 3\sqrt{\nu}A_n + cA_n^{3/2}.
    \]
    We claim that
    \[
      B_n\sqrt{2\nu n A_n} + \frac{B_n^2}{3}A_n
      \leq 2a_nD_n + D_n^2.
    \]
    Indeed, since \(A_n \geq 1\),
    \[
    \begin{aligned}
      B_n\sqrt{2A_n}
      =
      \roundb{\sqrt{\nu}+\sqrt{2\nu A_n}+cA_n}\sqrt{2A_n}
      =
      \sqrt{2\nu A_n}+2\sqrt{\nu}A_n+\sqrt{2}cA_n^{3/2}
      \leq
      2D_n.
    \end{aligned}
    \]
    Hence
    \[
      B_n\sqrt{2\nu n A_n} \leq 2a_nD_n.
    \]
    Similarly,
    \[
    \begin{aligned}
      B_n\sqrt{A_n/3}
      =
      \frac{1}{\sqrt{3}}
      \roundb{\sqrt{\nu A_n}+\sqrt{2\nu}A_n+cA_n^{3/2}}
      \leq
      D_n.
    \end{aligned}
    \]

    Combining these inequalities gives
    \[
      \sum_{t=1}^n X_t^2
      \leq
      a_n^2+2a_nD_n+D_n^2
      =
      (a_n+D_n)^2.
    \]
    Taking square roots yields
    \[
      \roundb[\bigg]{\sum_{t=1}^n X_t^2}^{1/2}
      \leq
      \sqrt{\nu n}
      +
      3\sqrt{\nu}A_n
      +
      cA_n^{3/2}.
    \]
    Recalling that \(A_n=\log(16n^2/\delta)\), and since \(n\geq 1\) was arbitrary and the event holds uniformly over \(n\), this proves the desired time-uniform bound.
\end{proof}
\clearpage

\section{Proof of lower bounds}\label{appendix:lower-bounds}

We prove the lower bounds stated in the main paper, restated here:

\lowerbound*

We first prove that the optimisation error bound implies the regret bound.

\begin{proof}[Proof of \cref{thm:lower-bound}, optimisation to regret reduction]
Fix a two-point online algorithm \(\cA\). From \(\cA\), construct a  two-point optimisation algorithm \(\bar{\cA}\) as follows. The
algorithm \(\bar{\cA}\) uses the same internal randomisation as \(\cA\), makes
the same two queries at each round, and observes the same feedback. If
\(x_1,\dots,x_n\in\Theta\) are the decisions made by \(\cA\), then
\(\bar{\cA}\) outputs
\[
    \hat{x}_n=\frac1n\sum_{t=1}^n x_t.
\]
Since \(\Theta\) is convex, \(\hat{x}_n\in\Theta\).

For every \(\mu\in\cM\), the function \(F_\mu\) is convex. Therefore, by
Jensen's inequality,
\[
    F_\mu(\hat{x}_n)
    \le
    \frac1n\sum_{t=1}^n F_\mu(x_t).
\]
Consequently,
\[
\begin{aligned}
    F_\mu(\hat{x}_n)-\inf_{u\in\Theta}F_\mu(u)
    &\le
    \frac1n\sum_{t=1}^n F_\mu(x_t)
    -
    \inf_{u\in\Theta}F_\mu(u) \\
    &=
    \frac1n
    \sup_{u\in\Theta}
    \sum_{t=1}^n
    \roundb[\big]{F_\mu(x_t)-F_\mu(u)}
    =
    \frac{R_n^\mu}{n}.
\end{aligned}
\]
Hence
\[
    R_n^\mu
    \ge
    n\roundb[\bigg]{
        F_\mu(\hat{x}_n)-\inf_{u\in\Theta}F_\mu(u)
    }.
\]
Applying \cref{thm:lower-bound} to the optimisation algorithm \(\bar{\cA}\)
gives
\[
    \sup_{\mu\in\cM}
    \mathbf P_\mu^\cA
    \curlyb[\bigg]{
        R_n^\mu
        \ge
        d^{(1/2+1/q-1/p)_+}
        \frac{LR\sqrt{n}}{256}
    }
    \ge
    \frac17.
\]
This proves the claim.
\end{proof}

We now turn to proving the optimisation error lower bound. For probability measures \(P\) and \(Q\), define the squared Hellinger distance
by
\[
    \Hh^2(P,Q)
    =
    \frac12\int
    \roundb[\Bigg]{
        \sqrt{\frac{\dif P}{\dif\lambda}}
        -
        \sqrt{\frac{\dif Q}{\dif\lambda}}
    }^2
    \dif\lambda,
\]
where \(\lambda\) is any measure dominating both \(P\) and \(Q\). Standard properties of the squared Hellinger and total variation distances used here are enumerated in Section 2.4 of \citet{tsybakov2009nonparametric}; beware, however, that \citet{tsybakov2009nonparametric} defines $\Hh^2$ without the factor of $1/2$.

The following is a sequential Hellinger chain result is the sequential version of the standard Hellinger product identity; see \citet[Section 2.4, Definition 2.3 and following properties]{tsybakov2009nonparametric}.

\begin{lemma}[Sequential Hellinger chain]\label{lem:lower-bound-hellinger-chain}
Let \(\cZ_0,\cY_1,\dots,\cY_n\) be measurable spaces and define
\[
    \cZ_t=\cZ_{t-1}\times\cY_t,
    \qquad t=1,\dots,n.
\]
Let \(P^0=Q^0\) be a probability measure on \(\cZ_0\). For each
\(t=1,\dots,n\), let \(K_t\) and \(L_t\) be probability kernels from
\(\cZ_{t-1}\) to \(\cY_t\), and define probability measures \(P^t,Q^t\) on
\(\cZ_t\) recursively by
\[
    P^t(\dif z,\dif y)=P^{t-1}(\dif z)K_t(z,\dif y),
    \qquad
    Q^t(\dif z,\dif y)=Q^{t-1}(\dif z)L_t(z,\dif y).
\]
Then
\[
    \Hh^2(P^n,Q^n)
    \le
    \frac12
    \sum_{t=1}^n
    \roundb[\bigg]{
        \int_{\cZ_{t-1}}\Hh^2(K_t,L_t)\dif P^{t-1}
        +
        \int_{\cZ_{t-1}}\Hh^2(K_t,L_t)\dif Q^{t-1}
    }.
\]
\end{lemma}

\begin{proof}[Proof of \cref{lem:lower-bound-hellinger-chain}]
For \(t=1,\dots,n\), the product formula for Hellinger integrals gives
\[
    1-\Hh^2(P^t,Q^t)
    =
    \int_{\cZ_{t-1}}
    \roundb[\big]{1-\Hh^2(K_t,L_t)}
    \sqrt{\dif P^{t-1}\dif Q^{t-1}}.
\]
Subtracting this identity from the corresponding identity for \(P^{t-1}\) and
\(Q^{t-1}\), we obtain
\[
    \Hh^2(P^t,Q^t)-\Hh^2(P^{t-1},Q^{t-1})
    =
    \int_{\cZ_{t-1}}
    \Hh^2(K_t,L_t)
    \sqrt{\dif P^{t-1}\dif Q^{t-1}}.
\]
Using \(2\sqrt{ab}\le a+b\) pointwise yields
\[
    \Hh^2(P^t,Q^t)-\Hh^2(P^{t-1},Q^{t-1})
    \le
    \frac12
    \roundb[\bigg]{
        \int_{\cZ_{t-1}}\Hh^2(K_t,L_t)\dif P^{t-1}
        +
        \int_{\cZ_{t-1}}\Hh^2(K_t,L_t)\dif Q^{t-1}
    }.
\]
Summing over \(t=1,\dots,n\), and using \(P^0=Q^0\), proves the claim.
\end{proof}

Now we turn to constructing our hard instances. We use the standard biweight kernel density \citep[page 3]{tsybakov2009nonparametric}. Let
\[
    \rho(u)=\frac{15}{16}(1-u^2)^2\1{u\in[-1,1]},
    \qquad u\in\R.
\]
For \(\ell\in\Np\), let \(U^{(\ell)}=(U_1,\dots,U_\ell)\), where
\(U_1,\dots,U_\ell\) are independent random variables with common density
\(\rho\).

\begin{lemma}[Product shift bound]\label{lem:lower-bound-product-shift}
Fix \(\ell\in\Np\). For \(c\in\R^\ell\), let \(\Pi_c^\ell\) denote the law of
\(U^{(\ell)}+c\). Then
\[
    \Hh^2(\Pi_c^\ell,\Pi_{-c}^\ell)\le 5\norm{c}_2^2.
\]
\end{lemma}

\begin{proof}
Let
\[
    \phi(u)=\sqrt{\rho(u)}=\frac{\sqrt{15}}4(1-u^2)_+.
\]
Since the coordinates of \(U^{(\ell)}\) are independent, the Hellinger
integral factorises. Hence
\[
    \Hh^2(\Pi_c^\ell,\Pi_{-c}^\ell)
    =
    1-\prod_{i=1}^\ell
    \roundb[\big]{1-\Hh^2(\pi_{c_i},\pi_{-c_i})}
    \le
    \sum_{i=1}^\ell\Hh^2(\pi_{c_i},\pi_{-c_i}),
\]
where \(\pi_a\) denotes the law of \(U_1+a\), for any $a\in \R$. For each
coordinate,
\[
    \Hh^2(\pi_{c_i},\pi_{-c_i})
    =
    \frac12
    \int_\R
    \roundb{\phi(u-c_i)-\phi(u+c_i)}^2
    \dif u.
\]
For almost every \(u\),
\[
    \phi(u+c_i)-\phi(u-c_i)
    =
    \int_{-1}^1 c_i\phi'(u+s c_i)\dif s.
\]
By Cauchy--Schwarz,
\[
    \roundb{\phi(u+c_i)-\phi(u-c_i)}^2
    \le
    2c_i^2\int_{-1}^1|\phi'(u+s c_i)|^2\dif s.
\]
Integrating over \(u\), multiplying by \(1/2\), and using Fubini and a change of
variables gives
\[
    \Hh^2(\pi_{c_i},\pi_{-c_i})
    \le
    2c_i^2\int_\R|\phi'(u)|^2\dif u.
\]
Since \(\phi'(u)=-(\sqrt{15}/2)u\) for \(u\in(-1,1)\), and
\(\phi'(u)=0\) almost everywhere outside \([-1,1]\),
\[
    \int_\R|\phi'(u)|^2\dif u
    =
    \frac{15}{4}\int_{-1}^1 u^2\dif u
    =
    \frac52.
\]
Thus \(\Hh^2(\pi_{c_i},\pi_{-c_i})\le5c_i^2\), and summing over
\(i=1,\dots,\ell\) proves the result.
\end{proof}

We next show that the amount of information provided by two queries to distinguish neighbouring hard instances is limited.

\begin{lemma}[Two-query information bound]\label{lem:lower-bound-two-query}
Fix \(\ell\in\Np\) and \(\alpha,\eta>0\). Let
\(\Sigma_\ell=\curlyb{-1,1}^\ell\). For \(\sigma\in\Sigma_\ell\),
\(Q\in\R^{2\times \ell}\) and \(m\in\R^2\), let \(\kappa_{Q,m}^\sigma\) denote
the law of
\[
    m+\alpha Q(\eta\sigma+U^{(\ell)}).
\]
For \(i=1,\dots,\ell\), let \(\sigma^{(i)}\) be obtained from \(\sigma\) by
flipping the \(i\)-th sign. Then, for every \(\sigma\in\Sigma_\ell\),
\[
    \sum_{i=1}^\ell
    \Hh^2\roundb[\big]{\kappa_{Q,m}^\sigma,\kappa_{Q,m}^{\sigma^{(i)}}}
    \le
    10\eta^2.
\]
\end{lemma}

\begin{proof}
Let \(v_i=Qe_i\). Since \(v_i\) is in the image of \(Q\), the equation
\(Qc=\eta\sigma_i v_i\) is solvable. Its minimum Euclidean norm solution is
\[
    c_i=\eta\sigma_i Q\tran(QQ\tran)^\dagger v_i,
\]
and
\[
    \norm{c_i}_2^2
    =
    \eta^2 v_i\tran(QQ\tran)^\dagger v_i.
\]
After subtracting \(m+\alpha Q(\eta\sigma-\eta\sigma_i e_i)\), the laws
\(\kappa_{Q,m}^\sigma\) and \(\kappa_{Q,m}^{\sigma^{(i)}}\) are the images,
under the same linear map \(u\mapsto \alpha Q u\), of the laws of
\(U^{(\ell)}+c_i\) and \(U^{(\ell)}-c_i\). By data processing for Hellinger
distance and \cref{lem:lower-bound-product-shift},
\[
    \Hh^2\roundb[\big]{\kappa_{Q,m}^\sigma,\kappa_{Q,m}^{\sigma^{(i)}}}
    \le
    5\norm{c_i}_2^2
    =
    5\eta^2 v_i\tran(QQ\tran)^\dagger v_i.
\]
Summing over \(i=1,\dots,\ell\) and using
\(\sum_{i=1}^\ell v_iv_i\tran=QQ\tran\), we obtain
\[
    \sum_{i=1}^\ell
    \Hh^2\roundb[\big]{\kappa_{Q,m}^\sigma,\kappa_{Q,m}^{\sigma^{(i)}}}
    \le
    5\eta^2
    \tr\roundb[\big]{(QQ\tran)^\dagger QQ\tran}
    =
    5\eta^2\rank(Q)
    \le
    10\eta^2. \qedhere
\]
\end{proof}

The following is a standard Assouad-type reduction, in a form convenient for an average edgewise total-variation bound; compare \citet[Lemma 2.12 and Theorem 2.12(ii)]{tsybakov2009nonparametric}.

\begin{lemma}[Assouad step]\label{lem:lower-bound-assouad}
Let \(\Sigma_s=\curlyb{-1,1}^s\), and let
\((P^\sigma)_{\sigma\in\Sigma_s}\) be probability measures on a measurable
space \((\cZ,\cG)\). For \(\sigma,\tau\in\Sigma_s\), write
$d_H(\sigma,\tau)=\sum_{i=1}^s\1{\sigma_i\ne\tau_i}$, and let
\(\sigma^{(i)}\) be obtained from \(\sigma\) by flipping the \(i\)-th sign.
Suppose that
\[
    \bar{T}
    =
    \frac{1}{2^s s}
    \sum_{\sigma\in\Sigma_s}
    \sum_{i=1}^s
    \TV(P^\sigma,P^{\sigma^{(i)}})
    \le
    \frac12.
\]
Then, for every measurable estimator \(\hat{\sigma}\colon\cZ\to\Sigma_s\),
there exists \(\sigma\in\Sigma_s\) such that
\[
    P^\sigma
    \curlyb[\bigg]{
        d_H(\hat{\sigma},\sigma)\ge \frac s8
    }
    \ge
    \frac17.
\]
\end{lemma}

\begin{proof}
For \(i=1,\dots,s\), define
\[
    M_i^+
    =
    2^{-(s-1)}
    \sum_{\mathclap{\substack{\sigma\in\Sigma_s\\ \sigma_i=1}}}P^\sigma,
    \qquad
    M_i^-
    =
    2^{-(s-1)}
    \sum_{\mathclap{\substack{\sigma\in\Sigma_s\\ \sigma_i=-1}}}P^\sigma.
\]
Let $A_i=\curlyb{z\in\cZ\colon \hat{\sigma}_i(z)=1}$. Then by the standard testing inequality \citep[e.g.][Theorem 2.2]{tsybakov2009nonparametric}
\[
    2^{-s}\sum_{\sigma\in\Sigma_s}
    P^\sigma\curlyb{\hat{\sigma}_i\ne\sigma_i}
    =
    \frac12\roundb[\big]{M_i^+(A_i^c)+M_i^-(A_i)}
    \ge
    \frac12\roundb[\big]{1-\TV(M_i^+,M_i^-)}.
\]
Moreover, by convexity of total variation,
\[
    \TV(M_i^+,M_i^-)
    \le
    2^{-(s-1)}
    \sum_{\mathclap{\substack{\sigma\in\Sigma_s\\ \sigma_i=1}}}
    \TV(P^\sigma,P^{\sigma^{(i)}}).
\]
Therefore,
\[
\begin{aligned}
    2^{-s}\sum_{\sigma\in\Sigma_s}
    \int d_H(\hat{\sigma}(z),\sigma)\dif P^\sigma(z)
    &=
    \sum_{i=1}^s
    2^{-s}\sum_{\sigma\in\Sigma_s}
    P^\sigma\curlyb{\hat{\sigma}_i\ne\sigma_i} \\
    &\ge
    \frac s2(1-\bar{T})
    \ge
    \frac s4.
\end{aligned}
\]
Hence there exists \(\sigma\in\Sigma_s\) such that
\[
    \int d_H(\hat{\sigma}(z),\sigma)\dif P^\sigma(z)\ge \frac s4.
\]
Since \(0\le d_H(\hat{\sigma},\sigma)\le s\), this implies
\[
    P^\sigma
    \curlyb[\bigg]{
        d_H(\hat{\sigma},\sigma)\ge \frac s8
    }
    \ge
    \frac17.
\]
Indeed, otherwise the last expectation would be strictly smaller than
\[
    \frac17s+\frac67\frac s8=\frac s4\,,
\]
which is impossible.
\end{proof}

We are now ready to prove the optimisation error bound of \cref{thm:lower-bound}.

\begin{proof}[Proof of \cref{thm:lower-bound}, optimisation error bound]
Fix a two-point algorithm \(\cA\). Let
\[
    a=\frac12+\frac1q-\frac1p,
\]
and fix \(s\in\curlyb{1,\dots,d}\), to be chosen at the end. Let
\[
    \pi_s(x)=(x_1,\dots,x_s)\in\R^s,
    \qquad x\in\Rd,
\]
and define
\begin{align}\label{eq:lower-bound-parameters}
    \eta_s=\frac1{16}\sqrt{\frac sn},
    \qquad
    \alpha_s=\frac{L}{2s^{1/q^*}},
    \qquad
    \tau_s=Rs^{-1/p},
    \qquad
    \beta_s=\alpha_s\eta_s.
\end{align}
Since \(s\le d\le n\), we have \(\eta_s\le1/16\). Define
\[
    \psi_{\tau_s}(u)=(|u|-\tau_s)_+,
    \qquad
    \Phi_s(x)=\beta_s\sum_{j=1}^s\psi_{\tau_s}(x_j),
    \qquad x\in\Rd.
\]
Let \(\Sigma_s=\curlyb{-1,1}^s\). For each \(\sigma\in\Sigma_s\), let
\(\mu^\sigma\) be the law of the random function
\[
    f^\sigma(x)=\Phi_s(x)+\alpha_s\ip{\eta_s\sigma+U^{(s)}}{\pi_s(x)},
    \qquad x\in\Rd.
\]

We first check that \(\mu^\sigma\in\cM\). Each realised function \(f^\sigma\) is
convex. Since \(\psi_{\tau_s}\) is \(1\)-Lipschitz on \(\R\),
\[
    |\Phi_s(x)-\Phi_s(y)|
    \le
    \beta_s\norm{\pi_s(x)-\pi_s(y)}_1
    \le
    \beta_s s^{1/q^*}\norm{x-y}_q
    =
    \frac{L\eta_s}{2}\norm{x-y}_q.
\]
Also, since \(U^{(s)}\in[-1,1]^s\) almost surely,
\[
    \alpha_s\norm{\eta_s\sigma+U^{(s)}}_{q^*}
    \le
    \alpha_s(1+\eta_s)s^{1/q^*}
    =
    \frac{L(1+\eta_s)}{2}.
\]
Since \(\eta_s\le1/16\), the sum of these two Lipschitz constants is at most
\(L\). Thus \(f^\sigma\) is \(L\)-Lipschitz with respect to \(\norm{\cdot}_q\).
The density \(\rho\) is symmetric, so \(U^{(s)}\) has mean zero. Therefore
\[
    F_{\mu^\sigma}(x)
    =
    \Phi_s(x)+\beta_s\ip{\sigma}{\pi_s(x)}.
\]
For brevity, write \(F_\sigma=F_{\mu^\sigma}\), and define
\(x_\sigma^\star\in\Rd\) by
\[
    x_{\sigma,j}^\star
    =
    \begin{cases}
        -\tau_s\sigma_j, & j\le s,\\
        0, & j>s.
    \end{cases}
\]
Then
\[
    \norm{x_\sigma^\star}_p=\tau_s s^{1/p}=R,
\]
with the usual convention when \(p=\infty\), and hence
\(x_\sigma^\star\in\Theta\). For \(x\in\Theta\), define
\[
    N_\sigma(x)=\sum_{j=1}^s\1{\sgn(x_j)\ne-\sigma_j}.
\]
Writing \(y_j=\sigma_jx_j\), we have
\[
    F_\sigma(x)-F_\sigma(x_\sigma^\star)
    =
    \beta_s\sum_{j=1}^s
    \roundb[\big]{\psi_{\tau_s}(y_j)+y_j+\tau_s}.
\]
For every \(y\in\R\), \(\psi_{\tau_s}(y)+y+\tau_s\ge0\). Moreover, if
\(\sgn(x_j)\ne-\sigma_j\), then \(\sigma_jx_j\ge0\), and hence
\[
    \psi_{\tau_s}(\sigma_jx_j)+\sigma_jx_j+\tau_s\ge \tau_s.
\]
Consequently, for every \(x\in\Theta\),
\[
    F_\sigma(x)-F_\sigma(x_\sigma^\star)
    \ge
    \beta_s\tau_s N_\sigma(x)
    =
    \alpha_s\eta_s R s^{-1/p}N_\sigma(x).
\]
In particular, \(x_\sigma^\star\) is a minimiser of \(F_\sigma\) over
\(\Theta\).

It remains to show that some sign vector cannot be identified accurately. We
make explicit the probability measures induced by the algorithm. Represent the
internal randomisation of \(\cA\) by a measurable space \(\cW\) with probability
law \(P_W\). Define
\[
    \cH_t=\cW\times(\R^2)^t,
    \qquad t=0,\dots,n.
\]
The algorithm is described by measurable query maps
\[
    b_t,c_t\colon\cH_{t-1}\to\Rd,
    \qquad t=1,\dots,n,
\]
and a measurable output map
\[
    x_n\colon\cH_n\to\Theta.
\]
For \(z\in\cH_{t-1}\), define
\[
    Q_t(z)=
    \begin{pmatrix}
        \pi_s(b_t(z))\tran\\
        \pi_s(c_t(z))\tran
    \end{pmatrix},
    \qquad
    m_t(z)=
    \begin{pmatrix}
        \Phi_s(b_t(z))\\
        \Phi_s(c_t(z))
    \end{pmatrix}.
\]
Let \(\Pi_0^s\) denote the law of \(U^{(s)}\). For \(\sigma\in\Sigma_s\), define
a probability kernel \(K_t^\sigma\) from \(\cH_{t-1}\) to \(\R^2\) by
\[
    K_t^\sigma(z,A)
    =
    \Pi_0^s
    \curlyb[\bigg]{
        u\in\R^s\colon
        m_t(z)+\alpha_s Q_t(z)(\eta_s\sigma+u)\in A
    },
    \qquad A\in\cB(\R^2).
\]
For each \(\sigma\in\Sigma_s\), define probability measures \(P^{\sigma,t}\) on
\(\cH_t\) recursively by
\[
    P^{\sigma,0}=P_W,
    \qquad
    P^{\sigma,t}(\dif z,\dif y)
    =
    P^{\sigma,t-1}(\dif z)K_t^\sigma(z,\dif y),
    \qquad t\in[n].
\]
Set
\[
    P^\sigma=P^{\sigma,n}.
\]
By construction, \(P^\sigma=\mathbf P_{\mu^\sigma}^{\cA}\).

For \(i=1,\dots,s\), let \(\sigma^{(i)}\) be obtained from \(\sigma\) by flipping
the \(i\)-th sign, and define
\[
    \Delta_{t,i}^\sigma(z)
    =
    \Hh^2\roundb[\big]{K_t^\sigma(z),K_t^{\sigma^{(i)}}(z)},
    \qquad z\in\cH_{t-1}.
\]
By \cref{lem:lower-bound-two-query}, for every \(z\in\cH_{t-1}\),
\[
    \sum_{i=1}^s\Delta_{t,i}^\sigma(z)\le 10\eta_s^2.
\]
Applying \cref{lem:lower-bound-hellinger-chain} to \(P^\sigma\) and
\(P^{\sigma^{(i)}}\) gives
\[
    \Hh^2(P^\sigma,P^{\sigma^{(i)}})
    \le
    \frac12
    \sum_{t=1}^n
    \roundb[\bigg]{
        \int_{\cH_{t-1}}\Delta_{t,i}^\sigma(z)\dif P^{\sigma,t-1}(z)
        +
        \int_{\cH_{t-1}}\Delta_{t,i}^\sigma(z)\dif P^{\sigma^{(i)},t-1}(z)
    }.
\]
Since
\[
    \Delta_{t,i}^{\sigma^{(i)}}(z)=\Delta_{t,i}^\sigma(z),
    \qquad z\in\cH_{t-1},
\]
the second term becomes the first term after reindexing
\(\sigma\mapsto\sigma^{(i)}\). Hence, averaging over
\(\sigma\in\Sigma_s\) and \(i=1,\dots,s\),
\[
\begin{aligned}
    \frac{1}{2^s s}
    \sum_{\sigma\in\Sigma_s}\sum_{i=1}^s
    \Hh^2(P^\sigma,P^{\sigma^{(i)}})
    \le
    \frac{1}{2^s s}
    \sum_{t=1}^n
    \sum_{\sigma\in\Sigma_s}
    \int_{\cH_{t-1}}
    \sum_{i=1}^s\Delta_{t,i}^\sigma(z)
    \dif P^{\sigma,t-1}(z)
    \le
    10\frac{n\eta_s^2}{s}.
\end{aligned}
\]
Using \(\TV^2(P,Q)\le2\Hh^2(P,Q)\) and Jensen's inequality,
\[
    \bar{T}
    :=
    \frac{1}{2^s s}
    \sum_{\sigma\in\Sigma_s}\sum_{i=1}^s
    \TV(P^\sigma,P^{\sigma^{(i)}})
    \le
    \sqrt{20\frac{n\eta_s^2}{s}}
    =
    \frac{\sqrt{20}}{16}
    <
    \frac12.
\]
Define \(\hat{\sigma}\colon\cH_n\to\Sigma_s\) by
\[
    \hat{\sigma}_j(z)=-\sgn(x_{n,j}(z)),
    \qquad j=1,\dots,s.
\]
By \cref{lem:lower-bound-assouad}, there exists \(\sigma\in\Sigma_s\) such that
\[
    P^\sigma
    \curlyb[\bigg]{
        d_H(\hat{\sigma},\sigma)\ge \frac s8
    }
    \ge
    \frac17.
\]
For this sign vector,
\[
    d_H(\hat{\sigma}(z),\sigma)=N_\sigma(x_n(z)).
\]
Therefore, on the event in the previous display,
\[
\begin{aligned}
    F_\sigma(x_n)-\inf_{x\in\Theta}F_\sigma(x)
    \ge
    \alpha_s\eta_s R s^{-1/p}\frac s8
    =
    \frac{LR}{256\sqrt{n}}s^{1/2+1/q-1/p}.
\end{aligned}
\]
Now choose \(s=d\) if \(a>0\), and \(s=1\) otherwise. Then
\[
    s^{1/2+1/q-1/p}=d^{(1/2+1/q-1/p)_+}.
\]
Since \(\mu^\sigma\in\cM\) and \(P^\sigma=\mathbf P_{\mu^\sigma}^{\cA}\), the
result follows.
\end{proof}

\clearpage

\end{document}